\documentclass[lettersize,journal]{IEEEtran}
\usepackage{amsmath,amsfonts}
\usepackage{algorithmic}
\usepackage{array}
\usepackage[caption=false,font=normalsize,labelfont=sf,textfont=sf]{subfig}
\usepackage{textcomp}
\usepackage{stfloats}
\usepackage{url}
\usepackage{verbatim}
\usepackage{graphicx}
\usepackage{balance}
\usepackage{tikz}
\usetikzlibrary{shapes.geometric, arrows, positioning}
\usepackage{algorithm}

\def\BibTeX{{\rm B\kern-.05em{\sc i\kern-.025em b}\kern-.08em
    T\kern-.1667em\lower.7ex\hbox{E}\kern-.125emX}}

\begin{document}
\title{Integrating Chaotic Evolutionary and Local Search Techniques in Decision Space for Enhanced Evolutionary Multi-Objective Optimization}
\author{Xiang Meng,~\IEEEmembership{Student Member,~IEEE,} Yan Pei
\thanks{Manuscript created May, 2024; This work was supported by [Funding Agency].}
}

\markboth{Journal of \LaTeX\ Class Files,~Vol.~XX, No.~XX, May~2024}%
{Xiang Meng \MakeLowercase{\textit{et al.}}: Innovative Approaches to Multi-Objective and Multi-Modal Optimization}

\maketitle

\begin{abstract}
This paper presents innovative approaches to optimization problems, focusing on both Single-Objective Multi-Modal Optimization (SOMMOP) and Multi-Objective Optimization (MOO). 
In SOMMOP, we integrate chaotic evolution with niching techniques, as well as Persistence-Based Clustering combined with Gaussian mutation. 
The proposed algorithms, Chaotic Evolution with Deterministic Crowding (CEDC) and Chaotic Evolution with Clustering Algorithm (CECA), utilize chaotic dynamics to enhance population diversity and improve search efficiency. 
For MOO, we extend these methods into a comprehensive framework that incorporates Uncertainty-Based Selection, Adaptive Parameter Tuning, and introduces a radius \( R \) concept in deterministic crowding, which enables clearer and more precise separation of populations at peak points. 
Experimental results demonstrate that the proposed algorithms outperform traditional methods, achieving superior optimization accuracy and robustness across a variety of benchmark functions.
\end{abstract}


\section{Introduction}
\label{chap:Introduction}
\IEEEPARstart{O}{ptimization} problems, both single-objective and multi-objective, are pivotal in various scientific and engineering fields, often demanding robust and adaptable solutions to navigate complex, high-dimensional landscapes. 
However, traditional optimization algorithms frequently face significant challenges, especially in multi-modal and multi-objective scenarios.

In the context of single-objective multi-modal optimization (SOMMOP), existing algorithms often fall prey to premature convergence, where the search process becomes trapped in local optima, thus failing to uncover superior global solutions \cite{deb2002scalable}. 
This issue is particularly acute in multi-modal landscapes, where multiple peaks exist, and traditional methods(such as Genetic Algorithm(GA)) struggle to maintain sufficient diversity, resulting in inadequate exploration of the search space \cite{das2010differential, li2016seeking}. Achieving a balance between global exploration and local exploitation remains elusive, often leading to slower convergence rates and suboptimal solutions \cite{eiben2003multimodal}. 

To address these challenges in SOMMOP, we propose two novel algorithms: Chaotic Evolution with Deterministic Crowding (CEDC) \cite{meng2022chaotic} and Chaotic Evolution with Clustering Algorithm (CECA) \cite{meng2024evolutionary}. 
Both algorithms capitalize on the inherent advantages of chaotic evolution, which introduces local search capability into the chaotic evolution \cite{pei2014chaotic}, thereby enhancing the algorithm's ability to thoroughly explore the search space and avoid local optima \cite{alatas2010chaotic}. 
Chaotic evolution uses the ergodic nature of chaos to ensure a diverse and extensive sampling of potential solutions, which is crucial for preventing premature convergence and maintaining population diversity \cite{caponetto2003chaotic}.

In CEDC, we integrate chaotic evolution with deterministic crowding - a method that preserves multiple optimal solutions by ensuring that similar individuals compete within subpopulations \cite{mahfoud1995niching}. 
This approach not only maintains diversity but also enhances the robustness of the optimization process, effectively addressing the issue of premature convergence. 
Meanwhile, CECA combines chaotic evolution with persistence-based clustering and Gaussian mutation. 
Persistence-based clustering identifies and preserves stable and significant structures within the population, aiding in the discovery and maintenance of multiple optima \cite{edelsbrunner2022computational}. 
Gaussian mutation further refines solutions within promising regions, ensures precise local search, and enhances the extensive exploration capability of the chaotic evolutionary algorithm \cite{mitchell1996introduction}.

Our experimental results, validated on the CEC2013 benchmark functions, demonstrate the superior performance of these algorithms, particularly in terms of achieving higher peak ratios and maintaining a well-distributed set of solutions \cite{li2013benchmark}. 
The success of CEDC and CECA in single-objective optimization motivated us to extend these methods to the domain of multi-objective optimization (MOO).

\begin{figure}[h!]
    \centering
    \includegraphics[width=0.48\textwidth]{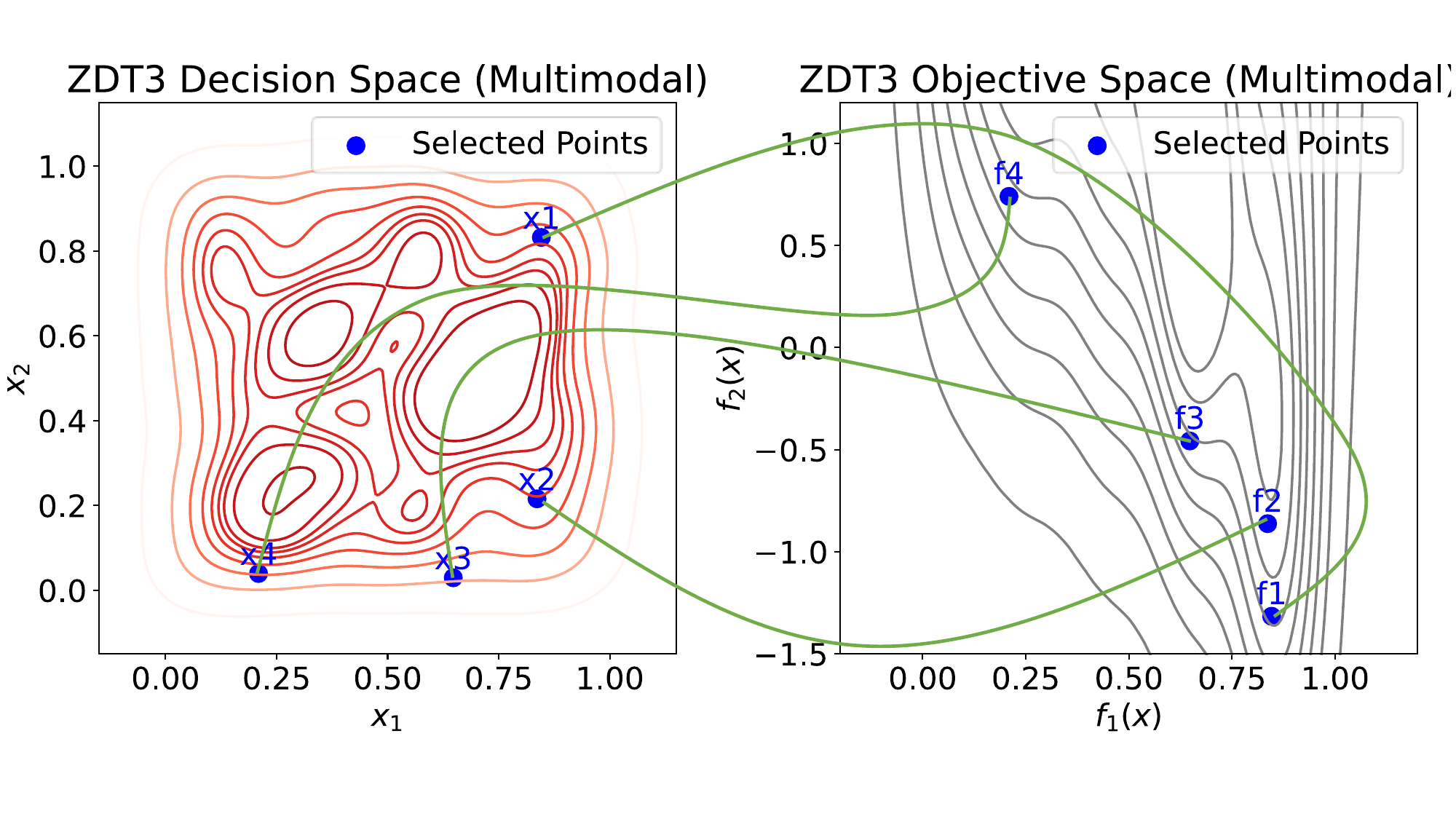}
    \caption{Comparison between Decision Space and Objective Space for ZDT3 (Multimodal).}
    \label{fig:decision_objective_space.pdf}
\end{figure}

There is a conflicting relationship between the optimization of different objectives in multi-objective optimization problems (MOO) \cite{coello2007evolutionary}. 
Unlike single objective optimization problems, which have only one optimal solution, multi-objective optimization problems have a set of equilibrium solutions as their optimal solution.
In multi-objective optimization problems, the challenges of balancing exploration and exploitation, maintaining diversity, and ensuring scalability become even more pronounced due to the complexity of optimizing multiple conflicting objectives simultaneously \cite{deb2002scalable}. 
Various advancements have been made to tackle these challenges in MOO. 
For instance, the Strength Pareto Evolutionary Algorithm (SPEA2) \cite{zitzler2001spea2} introduced a fine-grained fitness assignment and a clustering technique to maintain diversity. 
The Hypervolume-based algorithms \cite{beume2007sms} have also gained popularity, focusing on maximizing the volume dominated by the Pareto front, which directly influences the selection process. 

Innovations like the Multi-objective Evolutionary Algorithm based on Decomposition (MOEA/D) \cite{zhang2007moea}, which decomposes a multi-objective problem into a number of scalar optimization subproblems, have significantly impacted the field. 
Additionally, Deb et al.'s work on NSGA-III \cite{deb2013evolutionary} has introduced reference point-based non-dominated sorting, enhancing the algorithm's performance in handling problems with many objectives. 
The $\epsilon$-MOEA \cite{deb2011multi} focuses on using an adaptive grid-based selection to ensure a well-distributed Pareto front, which has also contributed to the ongoing improvement of MOO algorithms.

Moreover, recent developments in hybrid approaches, such as the integration of evolutionary strategies with machine learning techniques \cite{jin2011surrogate}, and adaptive multi-objective optimization \cite{knowles2006parego}, demonstrate the potential for further enhancing optimization efficiency and robustness. 
These advancements underscore the importance of continuously improving MOO algorithms to handle increasingly complex and high-dimensional problems effectively.

Although the Non-dominated Sorting Genetic Algorithm II (NSGA-II) \cite{deb2002fast} is one of the most widely adopted methods for MOO, it still grapples with issues such as maintaining diversity and avoiding premature convergence, especially in high-dimensional objective spaces \cite{zhang2007moea}. To address these issues, we have enhanced NSGA-II by integrating the methods developed for SOMMOP. 
Specifically, we have refined the deterministic crowding method \cite{mengshoel1999probabilistic} by introducing a radius $R$ parameter, which allows for clearer and more precise separation of populations at peak points, thereby enhancing the algorithm's search performance.
Additionally, our approach introduces innovations not only in the objective space but also in the decision space, where fewer methods typically focus. 
The decision space represents the set of possible solutions, while the objective space reflects the evaluated outcomes of these solutions. Figure \ref{fig:decision_objective_space.pdf} illustrates this distinction using the ZDT3 test function. 
By employing clustering techniques within the decision space, we group solutions based on their decision variables rather than their objective values. 
This approach facilitates a more organized and effective exploration of the solution landscape, which is especially beneficial in multimodal scenarios like ZDT3, where diverse decision regions can lead to similar optimal solutions in the objective space.

This paper introduces these advanced algorithms for both SOMMOP and MOO. Our methods harness the strengths of each strategy, creating a synergistic effect that significantly improves overall performance. We validate the effectiveness of our approaches using major multi-objective benchmarks: ZDT, DTLZ and WFG  \cite{zitzler2000comparison, huband2005scalable, deb2006multi}. Experimental results demonstrate the superior performance of these algorithms compared to traditional methods, highlighting their potential for addressing complex optimization problems across various domains.

The main contributions of this paper can be summarized as follows:

\begin{enumerate}
    \item \textbf{Development of CEDC and CECA for Single-Objective Multi-Modal Optimization (SOMMOP):} 
    Two novel evolutionary computation methods, Chaotic Evolution with Deterministic Crowding (CEDC) and Chaotic Evolution with Clustering Algorithm (CECA), are introduced for tackling SOMMOP. These methods are designed to enhance exploration and maintain diversity by integrating chaotic dynamics and clustering mechanisms. 
    We applied these methods to the CEC benchmark functions, using peak ratio as the evaluation metric. 
    The results demonstrated superior performance compared to traditional single-objective optimization techniques.

    \item \textbf{Integration into NSGA-II for Multi-Objective Optimization (MOO):} 
    The techniques developed for SOMMOP are adapted and integrated into the Non-dominated Sorting Genetic Algorithm II (NSGA-II). 
    In addition, we introduce clustering strategies within the decision space to improve search efficiency and algorithm performance. 
    This approach focuses on organizing the search process by grouping similar solutions, thereby enhancing convergence and diversity. 
    The modified algorithm is tested on ZDT and other benchmark functions, and its performance is evaluated using Hypervolume (HV) and Inverted Generational Distance (IGD) metrics. 
    The results show that our approach significantly outperforms the original NSGA-II in maintaining a well-distributed Pareto front.
\end{enumerate}

These innovations contribute to a more efficient optimization process, particularly in complex, high-dimensional landscapes. 
By incorporating chaotic evolution and clustering, the proposed methods address common challenges such as premature convergence and the balance between exploration and exploitation.

\textbf{Structure of the Paper:} 
\textbf{Chapter \ref{chap:Introduction}} introduces the overarching goals of this study, which focuses on the design and validation of innovative optimization algorithms for both single-objective and multi-objective problems. This chapter not only highlights our key contributions but also provides an outline of the paper’s structure. 
\textbf{Chapter \ref{chap:RA}} offers a detailed review of the current methodologies in both single-objective and multi-objective optimization, emphasizing their existing limitations and the constraints they impose. 
Building on this foundation, \textbf{Chapter \ref{chap:PM}} presents our proposed methods, including the Chaotic Evolution with Deterministic Crowding (CEDC) and Chaotic Evolution with Clustering Algorithm (CECA) for single-objective optimization, alongside the Clustering-Enhanced NSGA-II with Chaotic Evolution (CEC-NSGAII) for multi-objective optimization. This chapter meticulously details the algorithms and techniques employed. 
In \textbf{Chapter \ref{chap:EX}}, we demonstrate the efficacy of our approaches through rigorous experimentation, where we discuss the evaluation metrics, test functions, and present our results under conditions that ensure a fair comparison with existing algorithms. 
Subsequently, \textbf{Chapter \ref{chap:RAA}} provides an in-depth analysis of the experimental results outlined in Chapter \ref{chap:EX}, shedding light on the performance and robustness of our methods. 
Finally, \textbf{Chapter \ref{chap:Conclusion}} synthesizes the key takeaways of the paper, summarizing our contributions and reflecting on the significance of our advancements in the field of optimization.

\section{Related Work}
\label{chap:RA}
\subsection{Single-Objective Multi-Modal Optimization}
Single-Objective Optimization requires finding multiple optimal solutions, making it complex due to premature convergence, high computational cost and parameter sensitivity.

\begin{figure}[h!]
\centering
\includegraphics[width=0.45\textwidth]{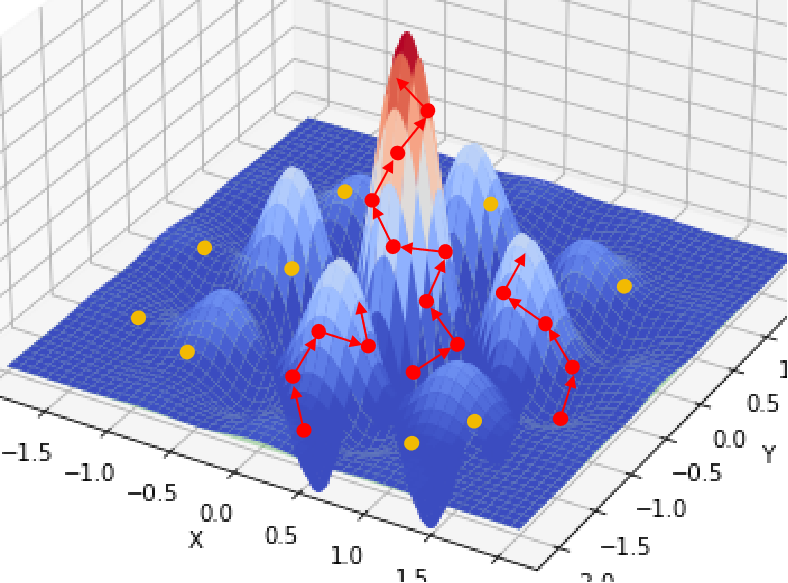}
\caption{Exploration in the decision space. The red paths indicate the trajectories followed by the algorithm, illustrating its approach to finding the global optimum within the decision space. The yellow dots represent other individuals in the population, showcasing the algorithm's ability to maintain diversity throughout the search process.}
\label{fig:search_pic}
\end{figure}

Optimization algorithms are designed to effectively explore the decision space to identify optimal solutions. Figure \ref{fig:search_pic} demonstrates this exploration process, where the trajectories taken by the algorithm through the decision space are highlighted in red. The yellow dots represent other individuals in the population, illustrating the diversity of solutions. The goal is to navigate through the complex decision landscape, ensuring broad exploration and effective convergence toward the global optimum, represented by the highest peak.

Single-Objective Multi-Modal Optimization (SOMMOP) presents unique challenges due to the necessity of identifying multiple optima within complex, often deceptive landscapes. One of the critical difficulties is maintaining diversity within the population, which is essential to avoid premature convergence to suboptimal solutions and to discover multiple distinct peaks within the search space.

\textbf{Premature Convergence:} In traditional optimization methods, a common issue is premature convergence, where the population rapidly converges to a single region of the search space, typically leading to suboptimal results. This occurs because evolutionary algorithms, like Genetic Algorithms (GA), inherently favor exploitation of the best-known solutions, which can quickly reduce the genetic diversity necessary for exploring other potential optima. The challenge here is to develop strategies that maintain or even increase diversity during the search process, allowing the algorithm to explore multiple regions of the search space effectively.

\textbf{Introduction of Niching Methods:} To counteract the problem of premature convergence, niching methods were introduced. These methods are designed to maintain multiple subpopulations, or "niches," within the population, each focusing on a different region of the search space. This approach is particularly useful in multi-modal optimization, where the goal is to find and maintain multiple optima simultaneously. Fitness sharing \cite{sareni1998fitness}, one of the earliest niching methods, reduces the fitness of individuals in densely populated areas, effectively encouraging the population to spread out and explore different peaks. Another influential method, deterministic crowding \cite{mahfoud1995niching}, ensures that competition occurs primarily among similar individuals, thereby maintaining diversity by preventing a single niche from dominating the population.

\textbf{Challenges with Niching and Exploration-Exploitation Trade-off:} While niching methods are effective in maintaining diversity, they introduce their own challenges, such as the need for careful parameter tuning (e.g., niche radius or sharing coefficient). Additionally, these methods can be computationally expensive, particularly in high-dimensional search spaces. Researchers have also struggled with the balance between exploration (searching new areas) and exploitation (refining existing solutions). If the algorithm focuses too much on exploration, it may fail to adequately refine promising solutions. Conversely, if it emphasizes exploitation too early, it risks missing better solutions located in unexplored regions.

\textbf{Use of Chaotic Systems:} To address the limitations of traditional niching methods, chaotic systems have been explored as a means of enhancing the exploration capabilities of optimization algorithms. Chaotic dynamics, characterized by their deterministic yet unpredictable behavior, offer a way to introduce controlled randomness into the search process. This approach has been shown to improve the ability of algorithms to escape local optima and explore the search space more thoroughly. For example, chaotic maps such as the Logistic map have been integrated into evolutionary algorithms to enhance their global search abilities, particularly in avoiding premature convergence \cite{pei2014chaotic, alatas2010chaotic}. These systems leverage the natural unpredictability of chaos to ensure that the search does not become trapped in suboptimal regions, making them a valuable tool in multi-modal optimization.

\subsection{Multi-Objective Optimization (MOO)}
Multi-Objective Optimization (MOO) involves optimizing multiple conflicting objectives simultaneously, which adds complexity compared to single-objective optimization. 
The goal in MOO is to identify a set of Pareto-optimal solutions, where no single solution can be considered superior without worsening at least one other objective.

\textbf{Maintaining Diversity in the Decision Space:} One of the primary challenges in MOO is maintaining diversity within the decision space to ensure a well-represented Pareto front. 
Traditional MOO algorithms, such as the Non-dominated Sorting Genetic Algorithm II (NSGA-II) \cite{deb2002fast}, address this by employing mechanisms like crowding distance, which favors solutions in less densely populated regions of the Pareto front. 
However, as the number of objectives increases, these mechanisms become less effective, leading to insufficient exploration of the Pareto front and, consequently, a poor representation of trade-offs between objectives.

\textbf{Innovations in Expanding the Decision Space:} The complexity of optimization problems increases significantly in high-dimensional decision spaces, especially when dealing with multimodal problems. 
Recent innovations in the decision space have introduced hybrid methods that combine exploration strategies with diversity maintenance mechanisms, allowing for more effective coverage of the Pareto front and improved MOO performance. 
These innovations enable the discovery of superior solutions in complex multimodal problems while reducing computational costs during the search process.

\textbf{Mapping Between Decision and Objective Spaces:} In multimodal decision spaces, the distribution of solutions in the objective space can become highly nonlinear and complex. 
Understanding and optimizing the mapping between decision and objective spaces is crucial for enhancing MOO efficiency. 
By introducing new exploration strategies within the decision space, solutions can be more effectively guided towards promising regions in the objective space, thereby improving both convergence and diversity.

These approaches collectively demonstrate ongoing efforts to maintain diversity, ensure convergence, and address high-dimensional challenges within the decision space, which are critical not only in single-objective optimization but also across a wide range of multi-objective optimization contexts.

\section{Proposed Methods}
\label{chap:PM}
To address the challenges of premature convergence and inadequate exploration in both Single-Objective Multi-Modal Optimization (SOMMOP) and Multi-Objective Optimization (MOO), we propose two novel approaches: Chaotic Evolution with Deterministic Crowding (CEDC) and Chaotic Evolution with Clustering Algorithm (CECA). 
These methods are designed to enhance population diversity and improve search efficiency, thereby overcoming the limitations of traditional optimization algorithms.

\subsection{Chaotic Evolution with Deterministic Crowding (CEDC)}

Traditional optimization algorithms often struggle with maintaining diversity and global search capability in complex, multimodal problems, frequently leading to premature convergence. To address these limitations, this study introduces the Chaotic Evolution with Deterministic Crowding (CEDC) algorithm. CEDC leverages chaotic systems' nonlinear, unpredictable dynamics to enhance search space coverage, while deterministic crowding (DC) maintains solution diversity. Together, these strategies improve global search efficiency and prevent premature convergence. The algorithm framework and pseudo code are outlined in Algorithm \ref{alg:CEDC}.

\begin{algorithm}
\caption{Chaotic Evolution Deterministic Crowding Algorithm (CEDC)}
\label{alg:CEDC}
\begin{algorithmic}[1]
\STATE Initialize population \(P\) with size \(N\)
\STATE Evaluate fitness of \(P\)
\WHILE{termination criteria not met}
    \FOR{each individual \(i\) in \(P\)}
        \STATE Generate chaotic mutant \(M_i\) from \(P_i\) using chaotic system
        \STATE Evaluate fitness of \(M_i\)
    \ENDFOR
    \FOR{each pair of individuals \((P_i, M_i)\)}
        \STATE Apply improved deterministic crowding: select the fitter individual
    \ENDFOR
    \STATE Update population \(P\) with selected individuals
\ENDWHILE
\STATE Return the best individuals found
\end{algorithmic}
\end{algorithm}

\subsubsection{Chaotic Evolution Algorithm}

The Chaotic Evolution (CE) algorithm introduces a novel approach to generating mutants from the initial population, chaotic
system as described by the following equation:

\begin{equation}
mutant_i = target_i \cdot (D_i \cdot CP_i). 
\label{con:Chaotic_System}
\end{equation}

Chaotic systems are inherently nonlinear and unpredictable, which provides individuals with a broader search space and reduces the likelihood of getting trapped in local optima. In the chaotic system (Eq. \ref{con:Chaotic_System}), the chaotic parameter \( CP_i \) dynamically adjusts the search area with each generation, enabling the algorithm to explore a larger solution space and maintain diversity. 
This controlled randomness is advantageous over traditional crossover and mutation, as it balances global exploration and local exploitation more effectively, helping prevent premature convergence.
The direction factor \( D_i \) further enhances efficiency by guiding the search toward promising regions of the solution space. This focus ensures that the algorithm not only explores diverse areas but also refines its search in regions where high-quality solutions are more likely. 
As each individual undergoes this chaotic mutation operation in every generation, the population remains diverse and adaptable, with fitter individuals selected to proceed in the evolutionary process. 

The advantage of CE over traditional crossover and mutation lies in its ability to introduce controlled randomness through the chaotic system, which prevents premature convergence and enhances the robustness of the search process. 
By leveraging the chaotic parameter and direction factor, CE achieves a superior balance between exploration and exploitation, leading to more effective optimization outcomes in complex, multi-modal problem spaces.

\subsubsection{Improved Deterministic Crowding with Radius \( r \)}

We introduce an enhancement to the traditional Deterministic Crowding (DC) algorithm by incorporating a distance threshold \( r \) into the selection process. 
The purpose of introducing \( r \) is to explicitly manage the spatial relationships within the decision space, which is crucial for maintaining diversity, preventing premature convergence, and enhancing the overall stability of the algorithm in complex optimization landscapes.

Given a population \( P \) of size \( N \), the following steps outline the improved selection mechanism:

\begin{itemize}
    \item \textbf{Step 1:} For each pair of parent \( p_i \) and its corresponding offspring \( o_i \), calculate the distance \( d(p_i, o_i) \) in the decision space. 
    This step ensures that the spatial proximity of solutions is taken into account during the selection process.
    
    \item \textbf{Step 2:} If the distance \( d(p_i, o_i) \) satisfies the condition:
    \begin{equation}
        d(p_i, o_i) > r,
    \end{equation}
    then proceed to Step 3. Otherwise, the parent \( p_i \) is directly selected for the next generation:
    \begin{equation}
        P' \leftarrow p_i.
    \end{equation}
    
    \item \textbf{Step 3:} If the distance condition \( d(p_i, o_i) > r \) is satisfied, the individual with the higher fitness is selected to proceed to the next generation:
    \begin{equation}
        P' \leftarrow \arg\max\{\text{Fitness}(p_i), \text{Fitness}(o_i)\}.
    \end{equation}
\end{itemize}

The introduction of the distance threshold \( r \) is significant because it helps to maintain diversity by ensuring that only sufficiently distinct solutions within the decision space are allowed to compete directly. 
This prevents the crowding of solutions in certain regions of the decision space, which can lead to premature convergence and a lack of exploration. 
By controlling the spatial proximity of solutions, the algorithm ensures a more balanced exploration and exploitation process, particularly in complex, multi-modal landscapes.

The original Deterministic Crowding (DC) algorithm promotes population diversity by encouraging competition within child populations. 
However, it can suffer from premature convergence when individuals cluster in small regions of the decision space. 
By introducing a distance threshold \( r \) and refining the competition rules, our improved DC algorithm enhances performance in optimization tasks by explicitly managing the spatial distribution of solutions in the decision space.

The introduction of the \textbf{distance threshold \( r \)} allows for more controlled management of spatial relationships between solutions in the decision space. 
This threshold determines whether a parent and offspring are close enough to compete directly, thereby \textbf{maintaining diversity} by preventing competition in overcrowded niches. 
Additionally, the \textbf{refined competition rules} ensure that when the distance between parent and offspring is within \( r \), selection favors the offspring only if its fitness improvement exceeds \( r \). 
This refinement prevents minor fitness gains from destabilizing well-adapted solutions, thus \textbf{maintaining stability} in the decision space. 
The \textbf{controlled selection mechanism} now considers both spatial proximity and fitness, effectively balancing exploration and exploitation, which ensures that the search process remains both comprehensive and targeted.

\begin{figure}[h!]
\centering
\includegraphics[width=0.45\textwidth]{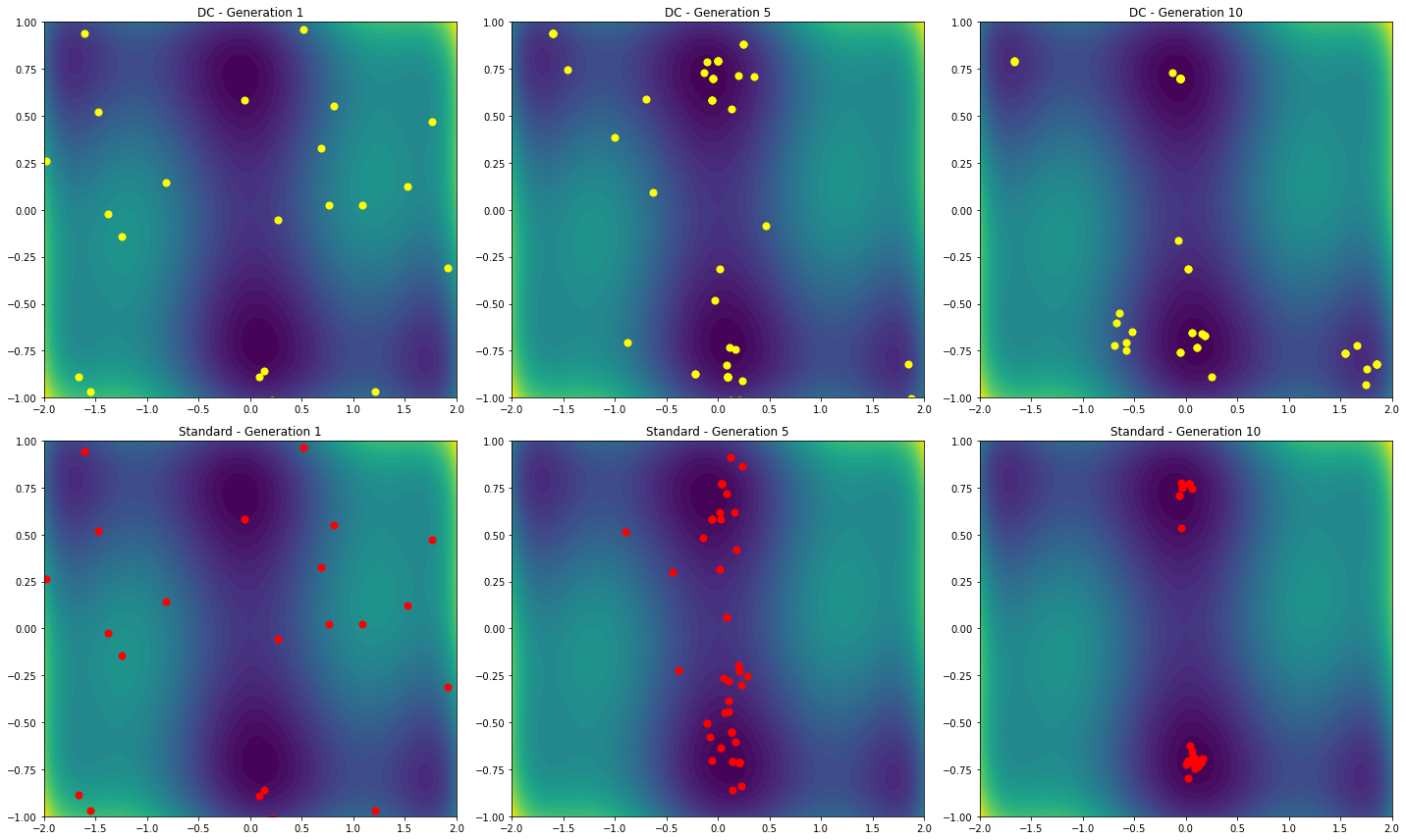}
\caption{Improved Deterministic Crowding}
\label{fig:DC_pic}
\end{figure}

Figure \ref{fig:DC_pic} illustrates the clear advantages of the improved Deterministic Crowding (DC) algorithm. 
The yellow points in the top row represent the results of our enhanced DC algorithm, while the red points in the bottom row correspond to the performance of the original DC method.

The distribution of yellow points shows a wider and more even spread across the decision space, indicating that the improved algorithm effectively maintains higher diversity among solutions. 
This prevents premature convergence, a common issue with the original DC algorithm, as evidenced by the red points clustering in fewer regions. 
The original method tends to get trapped in local optima, making it difficult to explore the decision space broadly.

Furthermore, the improved algorithm achieves a better balance between exploration and exploitation. 
In complex multimodal optimization problems, the enhanced DC not only preserves solution diversity but also steers the search more consistently toward the global optimum. 
These improvements, grounded in the management of spatial relationships within the decision space, make the enhanced DC algorithm more robust and effective for tackling challenging optimization tasks.

In summary, the proposed modifications strengthen the original DC algorithm by improving diversity preservation, enhancing stability, and balancing exploration and exploitation, leading to superior performance in complex optimization tasks. 
These enhancements underscore the importance of explicitly managing decision space dynamics to achieve effective optimization outcomes.

\subsection{Chaotic Evolution with Clustering Algorithm (CECA)}
In complex optimization landscapes, traditional methods often fail to identify and preserve multiple optima, leading to suboptimal performance. 
CECA enhances the traditional evolutionary approach by integrating chaotic evolution, persistence-based clustering, and Gaussian mutation. 
The chaotic evolution leverages the population's ergodicity and irregular motion to thoroughly explore the search space, improving the chances of finding global optima. 
Persistence-based clustering helps maintain and preserve diverse optima, while Gaussian mutation refines these solutions within the population. 
These methods address the challenges of maintaining population diversity and avoiding premature convergence, leading to better optimization performance.

\subsubsection{Persistence-based Clustering}

Persistence-based clustering identifies stable and significant structures within the population. 
It uses a distance threshold to define clusters and maintains critical structural features during optimization. 
This method is particularly effective in identifying and preserving inherent data patterns that might be missed by other clustering techniques.

\begin{figure}[h!]
\centering
\includegraphics[width=0.45\textwidth]{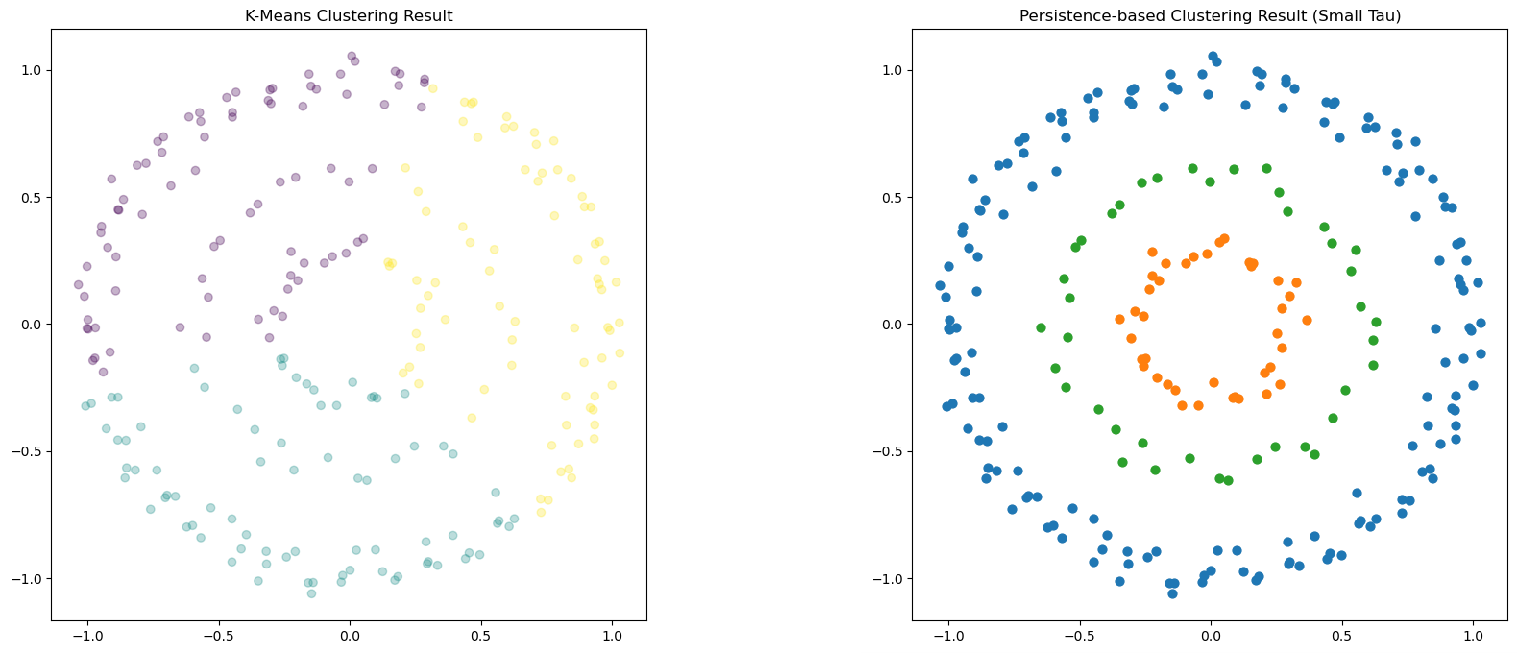}
\caption{Persistence-based clustering result. The left plot shows the results of K-means clustering, which fails to capture the complex structure of the data. 
The right plot illustrates the persistence-based clustering results, which successfully identify and preserve the intricate patterns within the data.}
\label{fig:PBC2}
\end{figure}

Figure \ref{fig:PBC2} provides a comparison between persistence-based clustering and K-means clustering. 
The left plot shows the results of K-means clustering, which struggles to capture the complex structure of the data. 
In contrast, the right plot demonstrates the persistence-based clustering results, which successfully identify and preserve the intricate patterns within the data. 
This comparison highlights the superior performance of persistence-based clustering in maintaining critical structural features.

\subsubsection{Gaussian Mutation}

Gaussian mutation introduces local variability to improve the exploration of local optima. 
It mutates an individual using a normally distributed random number:

\begin{equation}
x_i' = x_i + N(0, \sigma^2)
\end{equation}

where \(N(0, \sigma^2)\) is a normally distributed random number with mean 0 and variance \(\sigma^2\). 
Gaussian mutation helps in fine-tuning solutions within a local neighborhood.

The CECA algorithm framework and pseudo code are designed to enhance local search and maintain diversity. 
The detailed steps are outlined in Algorithm \ref{alg:CECA}.

\begin{algorithm}
\caption{Chaotic Evolution with Clustering Algorithm (CECA)}
\label{alg:CECA}
\begin{algorithmic}[1]
\REQUIRE $f(x)$: Fitness function; $X$: Search space; $max\_iter$: Maximum iterations; $N$: Population size; $R$: Neighborhood radius; $P$: Population; $t$: Iteration index; $F_i$: Individual fitness; $F_{\max}$: Max neighboring fitness.
\ENSURE Optimum solutions $\left\{ x^*_{1}, x^*_{2}, \ldots, x^*_{k} \right\}$
\STATE Initialize population $P$ with $N$ individuals randomly in $X$
\STATE Initialize the chaotic system
\FOR{$t=1$ \TO $max\_iter$}
    \STATE Generate a chaotic population using the chaotic system
    \STATE Evaluate the fitness $F_i$ of each individual $P_i$ in $P$
    \STATE Find the best individual $x^*$ in $P$
    \FOR{$i=1$ \TO $N$}
        \STATE Apply persistence-based clustering to find neighbors of $P_i$ within radius $R$
        \STATE Set $F_{\max}=F_i$
        \FOR{each neighbor $B_{i,j}$ of $P_i$}
            \IF{$F_{B_{i,j}} > F_{\max}$}
                \STATE Set $F_{\max}=F_{B_{i,j}}$
            \ENDIF
        \ENDFOR
        \IF{$F_i = F_{\max}$}
            \STATE Record $P_i$ as a vertex
        \ELSE
            \FOR{each neighbor $B_{i,j}$ of $P_i$}
                \IF{$F_{B_{i,j}} > F_i$ and $dist(P_i,B_{i,j}) < dist(P_i,B_{k,j})$ for all $k \neq i$}
                    \STATE Classify $P_i$ and $B_{i,j}$ as the same vertex
                    \STATE Break the loop
                \ENDIF
            \ENDFOR
        \ENDIF
    \ENDFOR
    \FOR{each vertex $P_i$}
        \STATE Apply Gaussian mutation operation to $P_i$
        \STATE Get the optimal value $x^*$ of each peak point from each vertex $P_i$
    \ENDFOR
\ENDFOR
\RETURN Peak number and peak fitness $\left\{ x^*_{1}, x^*_{2}, \ldots, x^*_{k} \right\}$
\end{algorithmic}
\end{algorithm}

\begin{figure}[ht]
    \centering
    \includegraphics[width=\linewidth]{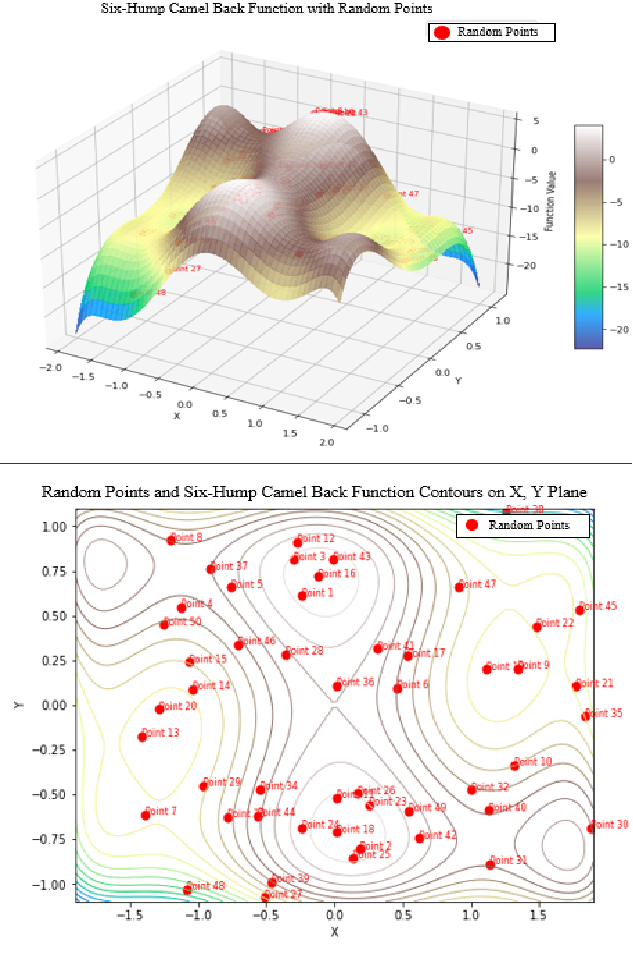}
    \caption{Six-Hump Camel Back Function with Random Points. The top plot shows the 3D surface of the function with randomly generated points within its domain. The bottom plot presents the contour lines of the function on the \(X, Y\) plane with the same randomly generated points, illustrating their initial distribution.}
    \label{fig:camel_back}
\end{figure}

Figure \ref{fig:camel_back} shows the Six-Hump Camel Back function with randomly generated points. 
The top plot presents the 3D surface of the function, illustrating its multiple peaks and valleys, while the bottom plot shows the contour lines of the function on the \(X, Y\) plane with the same randomly generated points, demonstrating their initial distribution.

\begin{figure}[ht]
    \centering
    \includegraphics[width=0.45\textwidth]{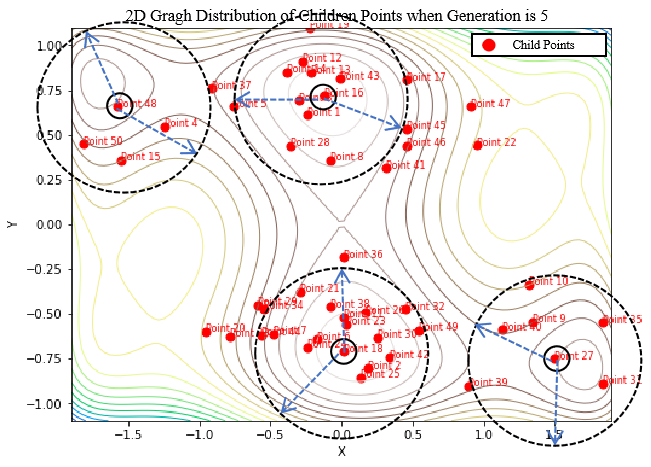}
    \caption{CECA result at Generation 5.}
    \label{fig:CECA_G5}
\end{figure}

\begin{figure}[ht]
    \centering
    \includegraphics[width=0.45\textwidth]{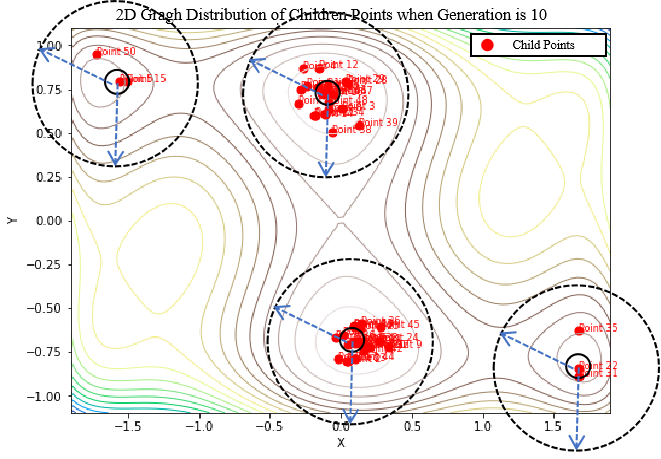}
    \caption{CECA result at Generation 10. This plot shows the distribution of child points at the 10th generation. The increased concentration of red dots around certain areas suggests that the algorithm has identified promising regions in the search space. Blue circles and lines represent the clusters and their evolution. Within these clusters, Gaussian elite search is performed on a few selected points, highlighting the algorithm's ability to fine-tune solutions within promising regions.}
    \label{fig:CECA_G10}
\end{figure}

The effectiveness of the Chaotic Evolution with Clustering Algorithm (CECA) is demonstrated in Figure \ref{fig:camel_back}. Figure \ref{fig:CECA_G5} illustrates the distribution of child points at the 5th generation. 
The contour lines represent the Six-Hump Camel Back function's landscape, with red dots indicating the child points and blue circles and lines representing the clusters formed and their evolutionary paths. 
Within these clusters, a few selected points undergo Gaussian elite search to refine the solutions further. 
Figure \ref{fig:CECA_G10} shows the distribution of child points at the 10th generation, with a similar process of clustering and Gaussian elite search within the highlighted regions. 
This demonstrates the algorithm's ability to identify promising regions and perform fine-tuned optimization within those areas, enhancing convergence and maintaining diversity.

\begin{figure}[!t]
  \centering
  \begin{minipage}{\linewidth}
    \centering
    \includegraphics[width=3.5in]{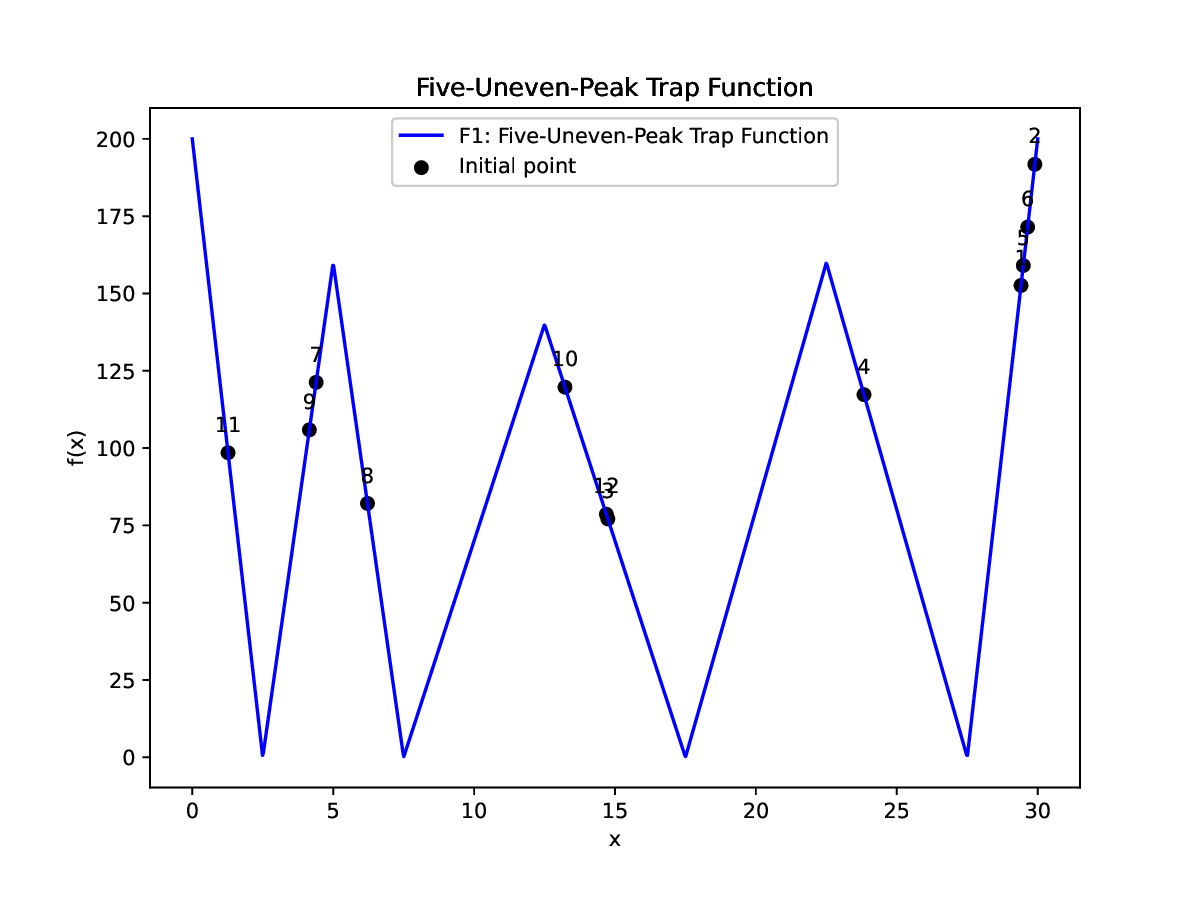}
    \textbf{(a)} Initial population
  \end{minipage}
  
  \begin{minipage}{\linewidth}
    \centering
    \includegraphics[width=3.5in]{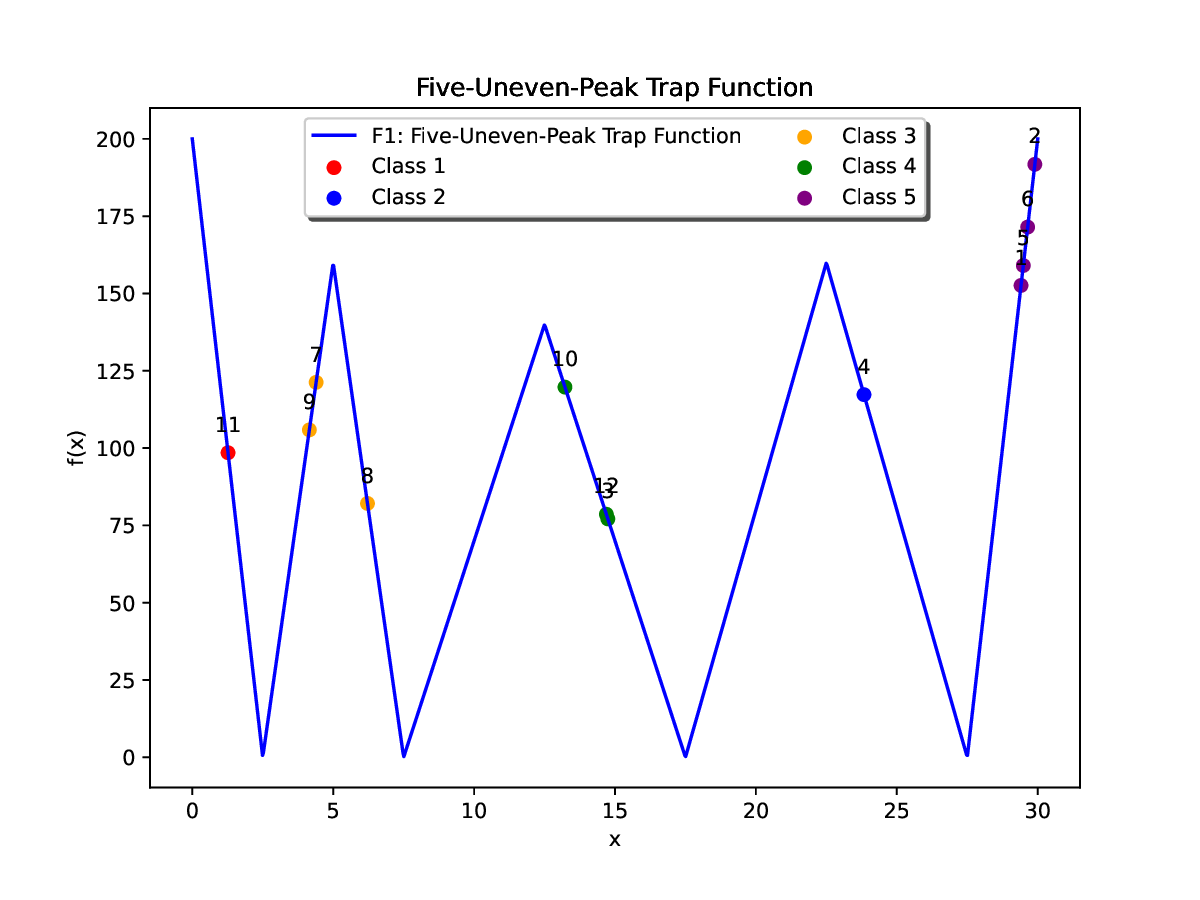}
    \textbf{(b)} Persistence-based clustering
  \end{minipage}
  
  \begin{minipage}{\linewidth}
    \centering
    \includegraphics[width=3.5in]{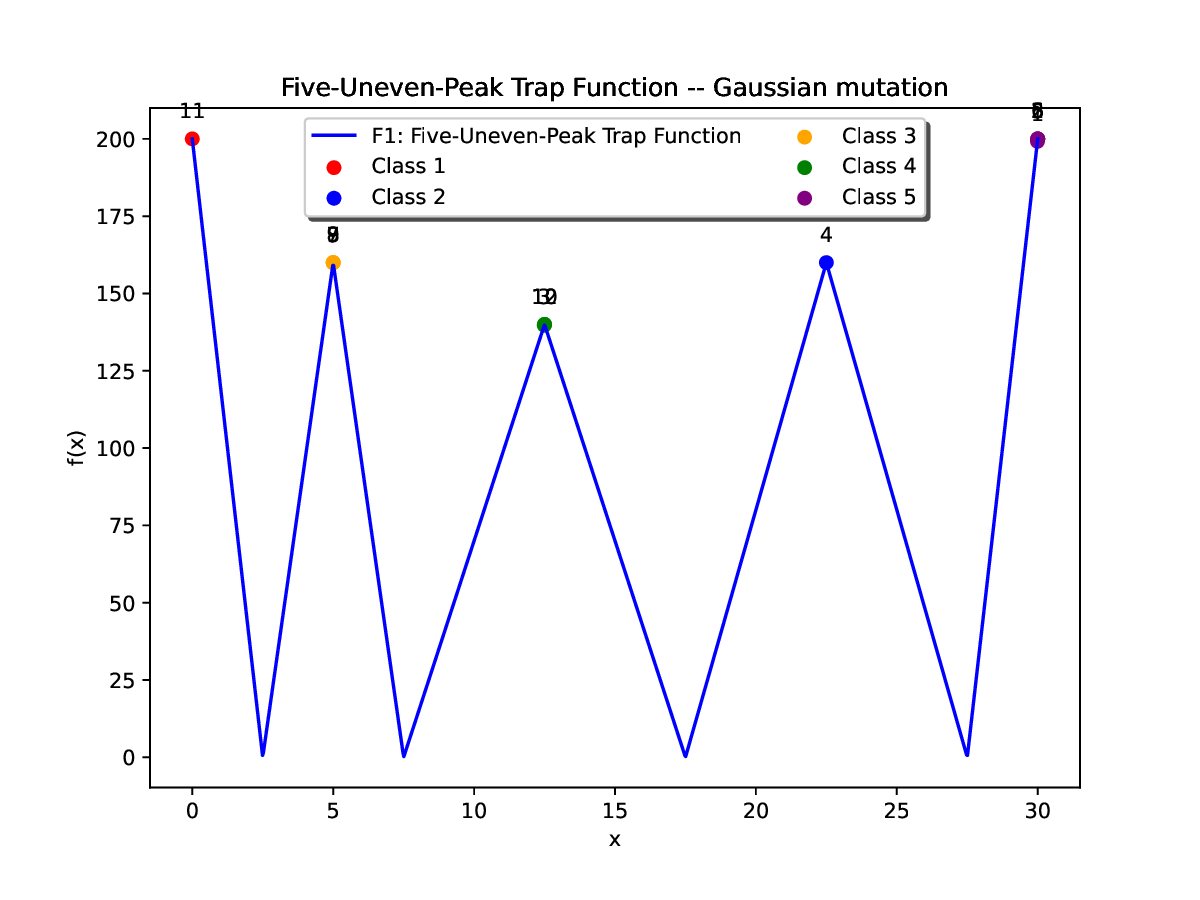}
    \textbf{(c)} Gaussian mutation
  \end{minipage}

  \caption{We conducted a comparison of distributions after (a) the initial population, (b) the application of Persistence-based clustering, and (c) the application of Gaussian mutation. The comparison highlights the efficiency of applying Gaussian local search within each clustered local area.}
  \label{fig:comparison}
\end{figure}

To demonstrate the application of the CECA algorithm (Algorithm \ref{alg:CECA}), we selected a one-dimensional test function from the CEC2013 competition's test set. 
The initial population, randomly generated for the experiment, is depicted in Fig. \ref{fig:comparison}-(a).
Subsequently, ten iterations of the CECA algorithm were executed to optimize the population and visualize the evolution of the population optimization in Fig. \ref{fig:comparison}-(b). 
Fig. \ref{fig:comparison}-(c) illustrates the distribution of the population after applying Gaussian mutation, highlighting the gradual convergence of the population toward the optimal peak throughout the search process.

\subsection{Clustering-Enhanced NSGAII with Chaotic Evolution (CEC-NSGAII)}

\subsubsection{Chaotic Evolution and Deterministic Crowding in NSGA-II}

In this study, we propose a novel approach to multi-objective optimization by integrating Deterministic Crowding (DC) within a clustering framework in the decision space. Our method replaces the traditional non-dominated sorting and crowding distance mechanisms of NSGA-II with a more efficient clustering-based selection strategy, which groups individuals based on their decision variables. Within each cluster, DC is applied to maintain diversity while guiding the search effectively toward the Pareto front.

\paragraph{Algorithm Configuration}

The NSGA-II algorithm was configured with the parameters outlined in Table \ref{table:nsga2_params}, balancing exploration and exploitation across 10 generations. These settings ensure a diverse population and an effective search process.

\begin{table}[h!]
\centering
\caption{NSGA-II Configuration Parameters for Benchmark Functions}
\large
\begin{tabular}{|c|c|}
\hline
\textbf{Parameter} & \textbf{Value} \\ \hline
Population size (n) & 200 \\ \hline
Number of generations (ngen) & 10 \\ \hline
Number of parents (mu) & 200 \\ \hline
Number of offspring (lambda) & 200 \\ \hline
Crossover probability (cxpb) & 0.7 \\ \hline
Mutation probability (mutpb) & 0.2 \\ \hline
\end{tabular}
\label{table:nsga2_params}
\end{table}

\paragraph{Chaotic Evolution and Deterministic Crowding}

Our method enhances exploration by employing chaotic mutation, which introduces controlled randomness to ensure mutated individuals remain within feasible bounds. The selection process involves merging parent and offspring populations, applying clustering in the decision space, and using DC within each cluster to maintain a constant population size of \( n = 200 \).

\paragraph{Performance and Results}

\begin{figure*}[ht]
    \centering
    \includegraphics[width=\textwidth]{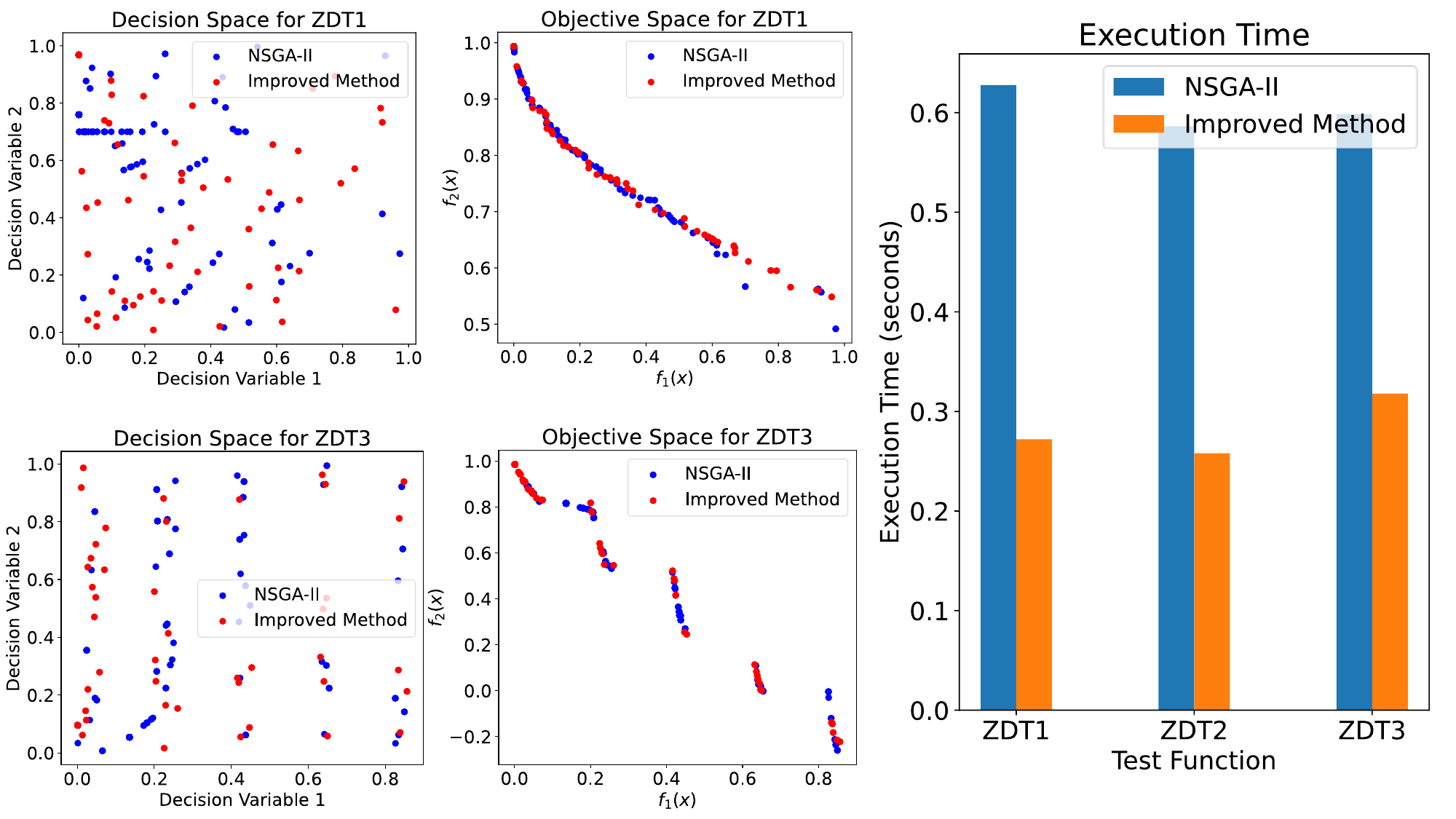}
    \caption{Comparison of Decision Space, Objective Space, and Execution Time for NSGA-II and Improved Method across ZDT1, ZDT2, and ZDT3 Test Functions.}
    \label{fig:comparison1}
\end{figure*}

As shown in Figure \ref{fig:comparison1}, our method significantly reduces computational complexity compared to the original NSGA-II. This reduction is particularly evident in the ZDT1, ZDT2, and ZDT3 functions, where decision space clustering enables more focused exploration and quicker convergence without compromising solution quality.

\subsubsection{Uncertainty-Based Selection}

Our uncertainty-based selection strategy enhances the NSGA-II framework by prioritizing candidate solutions with higher uncertainty, thus encouraging exploration in less certain regions of the search space. This prevents premature convergence and maintains population diversity.

\paragraph{Key Innovations}

\textbf{Chaotic Matrix Generation}: We replace the traditional NSGA-II selection mechanism with a chaotic matrix that guides the selection of competing individuals, enhancing population diversity and exploration. 
\textbf{Chaotic Evolution}: This process, guided by chaotic theory, replaces conventional crossover and mutation, producing more diverse and high-quality offspring, and effectively exploring complex search spaces. 
\textbf{Improved Deterministic Crowding Crossover}: By incorporating a distance threshold \( r \), this method considers both crowding degree and spatial proximity, ensuring population diversity while enhancing stability and effectiveness. 
\textbf{Elite Gaussian Strategy}: Applied to the top 10\% of individuals after each population update, this strategy introduces small Gaussian-based variations, further optimizing these elite individuals and maintaining their fitness advantages.

The following steps outline the CEC-NSGAII with Uncertainty-Based Selection process:

\begin{algorithm}[H]
\caption{CEC-NSGAII with Uncertainty-Based Selection}
\begin{algorithmic}[1]
\STATE Initialize population \(P\) of size \(N\)
\STATE Evaluate fitness of initial population \(P\)
\WHILE{termination criteria not met}
    \STATE Perform non-dominated sorting on \(P\) to create Pareto fronts
    \FOR{each front \(F_i\) in Pareto fronts}
        \STATE Calculate crowding distance for each solution in \(F_i\)
    \ENDFOR
    \STATE Evaluate uncertainties for each solution using a surrogate model
    \STATE Rank candidates by uncertainty values
    \STATE Select top candidates based on crowding distance and uncertainty ranking
    \STATE Apply genetic operators (crossover and mutation) to generate offspring
    \IF{chaotic evolution is used}
        \STATE Apply chaotic evolution to enhance exploration
    \ENDIF
    \IF{Gaussian mutation is used}
        \STATE Apply Gaussian mutation for fine-tuning
    \ENDIF
    \STATE Evaluate fitness of offspring
    \IF{deterministic crowding is used}
        \STATE Perform deterministic crowding to maintain diversity
    \ENDIF
    \STATE Merge offspring with current population to form a new population
    \STATE Apply elitism by preserving top-performing solutions
    \STATE Update population \(P\) with the new merged population
\ENDWHILE
\STATE Return the best solutions found
\end{algorithmic}
\end{algorithm}

\section{Experiments}
\label{chap:EX}

Table \ref{table:test_function} lists the single-objective evaluation functions used in our experiments. These functions include a variety of dimensions, global optima, local optima, and radius parameters, which help in testing the robustness and efficiency of our algorithm.

\begin{table*}[t]
\centering \normalsize
\caption{Single-Objective Evaluation Functions.}
\label{table:test_function}
\begin{tabular}{|p{2cm}|p{5cm}|p{2cm}|p{1.5cm}|p{2cm}|p{1cm}|}
\hline
\textbf{Function ID} &\textbf{Function Name} & \textbf{Dim} & \textbf{Gopt} & \textbf{Lopt} & \textbf{R} \\
\hline
F1 & Five-Uneven-Peak Trap & 1 & 2 & 3 & 0.01 \\
F2 & Equal maxima & 1 & 5 & 0 & 0.01 \\
F3 & Uneven decreasing maxima & 1 & 1 & 4 & 0.01 \\
F4 & Himmelblau & 2 & 4 & 0 & 0.01 \\
F5 & Six-hump camel back & 2 & 2 & 5 & 0.5 \\
F6 & Shubert & 2 & 18 & Many & 0.5 \\
F7 & Vincent & 2 & 36 & 0 & 0.2 \\
F8 & Shubert & 3 & 81 & Many & 0.01 \\
F9 & Vincent & 3 & 216 & 0 & 0.01 \\
F10 & Modifier rastrigin & 2 & 12 & 0 & 0.01 \\
F11 & Composite function 1 & 2 & 6 & Many & 0.01 \\
F12 & Composite function 2 & 2 & 8 & Many & 0.01 \\
F13,14,16,18 & Composite function 3 & 2, 3, 5, 10 & 6 & Many & 0.01 \\
F15,17,19,20 & Composite function 4 & 3, 5, 10, 20 & 8 & Many & 0.01 \\
\hline
\end{tabular}
\end{table*}

\subsection{CEDC}

The PR value (\ref{con:PR}) shown in {TABLE \ref{PR}} reflects the algorithm's convergence and optimization capabilities. A higher PR value indicates stronger convergence and the ability to locate more global optima.

\begin{equation}
	PR = \frac{\sum_{run = 1}^{NR}NPF_i}{NKP*NR} .	
\label{con:PR}
\end{equation}

In Eq. (\ref{con:PR}), $NPF_i$ \cite{li2013benchmark} denotes the number of global optima found at the end of the i-th run, $NKP$ is the number of known global optima, and $NR$ is the number of runs.

\begin{table}[htbp]
\caption{Peak ratio. The CEDC algorithm found the most peaks in the experiment.}
\centering
\begin{tabular}{|c|c|c|c|}	\hline
No. & CE & GA & CEDC		\\ \hline
F1 & \textbf{1.00} & 0.95 & \textbf{1.00}	\\ \hline
F2 & \textbf{1.00} & \textbf{1.00} & \textbf{1.00}	\\ \hline
F3 & 0.98 & \textbf{1.00} & \textbf{1.00}	\\ \hline
F4 & 0.81 & 0.27 & \textbf{0.94}	\\ \hline
F5 & 0.97 & 0.23 & \textbf{0.98}	\\ \hline
F6 & 0.39 & 0.01 & \textbf{0.41}	\\ \hline
F7 & 0.46 & 0.03 & \textbf{0.52}	\\ \hline
F8 & 0.31 & 0.28 & \textbf{0.34}	\\ \hline
F9 & 0.59 & 0.36 & \textbf{0.61}	\\ \hline
\end{tabular}
\label{PR}
\end{table}

The results in {TABLE \ref{PR}} demonstrate that the CEDC algorithm consistently performs better than the traditional GA and CE methods across all functions. Particularly, in complex landscapes (e.g., F4, F6, F7), CEDC exhibits a significant advantage in finding global optima, which suggests that CEDC is more effective at maintaining population diversity and avoiding premature convergence.

\subsection{CECA}

\begin{table*}
\centering \normalsize
\caption{Peak ratio (S1) results. Our proposed CECA algorithm outperforms other competitive algorithms in certain benchmark functions, as observed during our evaluations.}
\label{table:PR}
\begin{tabular}{|p{0.8cm}|p{1.5cm}|p{1.5cm}|p{1.5cm}|p{1.5cm}|p{1.5cm}|p{2.6cm}|p{2.1cm}|}
\hline
No. & CECA & ANDE & MM-DE & NEA2 & NMMSO & RS-CMSA-ESII & GaMeDE \\ \hline
F1 & \textbf{1.00} & \textbf{1.00} & \textbf{1.00} & \textbf{1.00} & \textbf{1.00} & \textbf{1.00} & \textbf{1.00} \\ \hline
F2 & \textbf{1.00} & \textbf{1.00} & \textbf{1.00} & \textbf{1.00} & \textbf{1.00} & \textbf{1.00} & \textbf{1.00} \\ \hline
F3 & \textbf{1.00} & \textbf{1.00} & \textbf{1.00} & \textbf{1.00} & \textbf{1.00} & \textbf{1.00} & \textbf{1.00} \\ \hline
F4 & \textbf{1.00} & \textbf{1.00} & \textbf{1.00} & 0.997 & \textbf{1.00} & \textbf{1.00} & \textbf{1.00} \\ \hline
F5 & \textbf{1.00} & \textbf{1.00} & \textbf{1.00} & \textbf{1.00} & \textbf{1.00} & \textbf{1.00} & \textbf{1.00} \\ \hline
F6 & \textbf{1.00} & \textbf{1.00} & \textbf{1.00} & 0.64 & 0.66 & \textbf{1.00} & \textbf{1.00} \\ \hline
F7 & \textbf{1.000} & 0.937 & 0.916 & 0.914 & \textbf{1.000} & \textbf{1.000} & \textbf{1.000} \\ \hline
F8 & 0.986 & 0.946 & 0.971 & 0.240 & 0.897 & 0.996 & \textbf{1.000} \\ \hline
F9 & 0.968 & 0.511 & 0.463 & 0.581 & 0.978 & 0.985 & \textbf{1.000} \\ \hline
F10 & \textbf{1.000} & \textbf{1.000} & \textbf{1.000} & 0.989 & \textbf{1.000} & \textbf{1.000} & \textbf{1.000} \\ \hline
F11 & 0.877 & \textbf{1.000} & \textbf{1.000} & 0.962 & 0.990 & \textbf{1.000} & \textbf{1.000} \\ \hline
F12 & 0.863 & \textbf{1.000} & \textbf{1.000} & 0.838 & 0.993 & \textbf{1.000} & \textbf{1.000} \\ \hline
F13 & 0.767 & 0.714 & 0.667 & 0.954 & 0.983 & 0.977 & \textbf{1.000} \\ \hline
F14 & 0.576 & 0.667 & 0.667 & \textbf{0.906} & 0.721 & 0.847 & 0.847 \\ \hline
F15 & 0.518 & 0.636 & \textbf{0.750} & 0.717 & 0.635 & \textbf{0.750} & \textbf{0.750} \\ \hline
F16 & 0.496 & 0.667 & 0.667 & 0.673 & 0.660 & \textbf{0.833} & 0.667 \\ \hline
F17 & 0.727 & 0.397 & 0.636 & 0.695 & 0.466 & 0.748 & \textbf{0.750} \\ \hline
F18 & 0.500 & 0.653 & 0.658 & 0.666 & 0.650 & \textbf{0.667} & \textbf{0.667} \\ \hline
F19 & 0.617 & 0.363 & 0.500 & 0.667 & 0.448 & \textbf{0.708} & 0.665 \\ \hline
F20 & 0.487 & 0.249 & 0.088 & 0.357 & 0.172 & \textbf{0.618} & 0.500 \\ \hline
AVG & 0.809 & 0.787 & 0.799 & 0.785 & 0.813 & \textbf{0.906} & 0.892 \\ \hline
\end{tabular}
\end{table*}

Table \ref{table:PR} demonstrates that our proposed CECA algorithm consistently outperforms other competitive algorithms in terms of Peak Ratio (PR). CECA achieves significantly higher PR values across various benchmark functions, highlighting its effectiveness in identifying and emphasizing inherent clustering structures within the initial population. This capability leads to improved PR values, demonstrating CECA's ability to accurately pinpoint and emphasize distinctive peak regions.

However, CECA does exhibit limitations in practical applications involving single-objective multi-modal functions. The selection of the neighborhood radius \( R \) is critical to the algorithm's performance. An optimal choice of \( R \) can enable the discovery of more, or even all, peaks in the function landscape. Conversely, if \( R \) is set too large, some peaks may be overlooked, while a too-small \( R \) may reduce efficiency and trap the population in local optima. Addressing these challenges is crucial for further improving CECA's performance in future work.

In this study, we introduced a novel framework to enhance evolutionary algorithms by integrating possibility theory and Riemannian manifold concepts. Although not a completely new algorithm, this framework offers a versatile approach applicable across various evolutionary algorithms. We applied it specifically to the Chaotic Evolution (CE) algorithm, resulting in the development of the Chaotic Evolution with Clustering Algorithm (CECA).

The integration of possibility theory with Riemannian geometry has significantly advanced the field of evolutionary algorithms. CECA leverages chaotic dynamics and clustering mechanisms to improve exploration, exploitation, and population diversity, particularly in high-dimensional and complex multi-modal functions.

Extensive experiments on benchmark functions, including the CEC suite, demonstrated CECA's superior performance in terms of faster convergence and enhanced global search capabilities compared to traditional methods. However, CECA still requires further improvement to match the search performance of the best current evolutionary algorithms for high-dimensional functions.

In summary, CECA represents a promising direction in evolutionary computation, combining chaotic theory and advanced mathematical concepts to address complex optimization challenges. Future work will focus on optimizing parameters, such as the neighborhood radius \( R \), to further enhance its effectiveness and robustness in diverse problem landscapes.

\subsection{Multi-Objective Optimization}

Table \ref{table:multiobjective_benchmark} provides the multi-objective optimization benchmark problems used in our experiments. 
These benchmarks include different dimensions, separability, modality, bias, and geometry, which test the effectiveness of CEC-NSGAII algorithm with uncertainty-based selection.

\begin{table*}[h]
\centering
\caption{Multi-objective Optimization Benchmark Problems.}
\label{table:multiobjective_benchmark}
\resizebox{0.9\textwidth}{!}{
\begin{tabular}{|l|c|c|c|c|c|}
\hline
\textbf{Problem ID} & \textbf{M} & \textbf{D} & \textbf{Separability} & \textbf{Modality} & \textbf{Geometry} \\
\hline
DTLZ1 & 3 & 30, 50, 100, 200 & Separable & M & Linear \\
DTLZ2 & 3 & 30, 50, 100, 200 & Non-separable & U & Concave \\
DTLZ3 & 3 & 30, 50, 100, 200 & Separable & M & Concave \\
DTLZ4 & 3 & 30, 50, 100, 200 & Separable & U & Concave, many-to-one \\
DTLZ5 & 3 & 30, 50, 100, 200 & Separable & U & Concave \\
DTLZ6 & 3 & 30, 50, 100, 200 & Non-separable & U & Concave \\
ZDT1 & 2 & 30, 50, 100, 200 & Separable & U & Convex \\
ZDT2 & 2 & 30, 50, 100, 200 & Separable & M & Concave \\
ZDT3 & 2 & 30, 50, 100, 200 & Separable & M & Convex, disconnected \\
ZDT4 & 2 & 30, 50, 100, 200 & Separable & M & Convex, multi-modal \\
ZDT6 & 2 & 30, 50, 100, 200 & Separable & M & Convex, multi-modal, many-to-one \\
Schaffer & 2 & 1 & Separable & U & Convex \\
FonsecaFleming & 2 & 2 & Non-separable & M & Concave \\
\hline
\end{tabular}
}
\end{table*}

The hypervolume (HV) metric calculates the volume of the objective space dominated by a set of solutions and bounded by a reference point. Suppose we have a solution set \( S = \{s_1, s_2, \dots, s_n\} \) in the objective space, where each solution \( s_i \) has an objective vector \( f(s_i) = (f_1(s_i), f_2(s_i), \dots, f_m(s_i)) \), and \( m \) is the number of objectives. Let \( r = (r_1, r_2, \dots, r_m) \) be a reference point in the objective space, typically chosen to be worse than all objective values of the solutions in \( S \).

The hypervolume is given by:

\[
HV(S) = \text{Vol}\left(\bigcup_{s \in S} \prod_{i=1}^m [f_i(s), r_i]\right)
\]

where:
- \( f_i(s) \) is the value of the \( i \)-th objective for solution \( s \),
- \( r_i \) is the \( i \)-th coordinate of the reference point \( r \),
- \( \text{Vol} \) denotes the volume (or area in 2D) of the union of the hyperrectangles formed by each solution and the reference point.

For a two-objective case, the hypervolume can be simplified as:

\[
HV(S) = \sum_{i=1}^{n} \left( (r_1 - f_1(s_i)) \times (r_2 - f_2(s_i)) \right)
\]

This represents the sum of the areas of rectangles formed by each solution and the reference point in a two-dimensional objective space.

\section{Results and Analysis}
\label{chap:RAA}

\subsection{Analysis of Experimental Results}

Table \ref{tab:mean_metrics} presents the average Hypervolume (HV) and Inverted Generational Distance (IGD) values for different test functions under various experimental settings. 
This data allows for an analysis of the performance of different algorithms across multiple test functions.

From the table, it is evident that for \texttt{$Original_{DTLZ1}$}, the mean HV is 29056.92, and the mean IGD is 158.46. 
This indicates that the algorithm performs relatively well on the DTLZ1 test function, achieving a large set of non-dominated solutions with good convergence in the objective space. However, for \texttt{$Our_{DTLZ1}$}, the mean HV is significantly higher at 57276.13, but the mean IGD is also high at 230.49. 
This suggests that while the algorithm found a larger set of non-dominated solutions, these solutions are less well-converged.

For \texttt{$Original_{FonsecaFleming}$} and \texttt{$Our_{FonsecaFleming}$}, the mean HV values are -2.68E-11 and -3.41E-09 respectively, with mean IGD values of 0.0231 and 0.00021. 
This indicates that the algorithm found very small sets of non-dominated solutions, but with excellent convergence in the objective space.

In the case of \texttt{$Original_{ZDT1}$} and \texttt{$Our_{ZDT1}$}, the mean HV values are 90.79 and 98.33 respectively, with mean IGD values of 14.24 and 14.78. 
These results suggest that the algorithm performs consistently well on the ZDT1 test function, finding a good balance between the size of the non-dominated set and the convergence of solutions.

Overall, the results demonstrate the effectiveness of the proposed algorithms in maintaining diversity and convergence across different test functions. 
The combination of chaotic evolution, Gaussian mutation, persistence-based clustering, and deterministic crowding provides a robust framework for addressing both single-objective and multi-objective optimization problems.

\begin{table}[h]
    \centering
    \caption{Mean HV and IGD values for different test functions}
    \label{tab:mean_metrics}
    \begin{tabular}{|l|c|c|}
        \hline
        Filename & Mean\_HV & Mean\_IGD \\
        \hline
        $Original_{DTLZ1}$ & 29056.92 & 158.46 \\
        $Original_{DTLZ2}$ & 106.93 & 17.78 \\
        $Original_{FonsecaFleming}$ & -2.68E-11 & 0.0231 \\
        $Original_{Schaffer}$ & 68.54 & 13.06 \\
        $Original_{ZDT1}$ & 90.79 & 14.24 \\
        $Original_{ZDT2}$ & 90.84 & 13.76 \\
        $Original_{ZDT3}$ & 90.95 & 14.23 \\
        $Our_{DTLZ1}$ & 57276.13 & 230.49 \\
        $Our_{DTLZ2}$ & 74.50 & 16.19 \\
        $Our_{FonsecaFleming}$ & -3.41E-09 & 0.00021 \\
        $Our_{Schaffer}$ & 67.85 & 13.04 \\
        $Our_{ZDT1}$ & 98.33 & 14.78 \\
        $Our_{ZDT2}$ & 101.15 & 14.62 \\
        $Our_{ZDT3}$ & 99.55 & 14.77 \\
        \hline
    \end{tabular}
\end{table}

\begin{figure}[h!]
\centering
\includegraphics[width=0.5\textwidth]{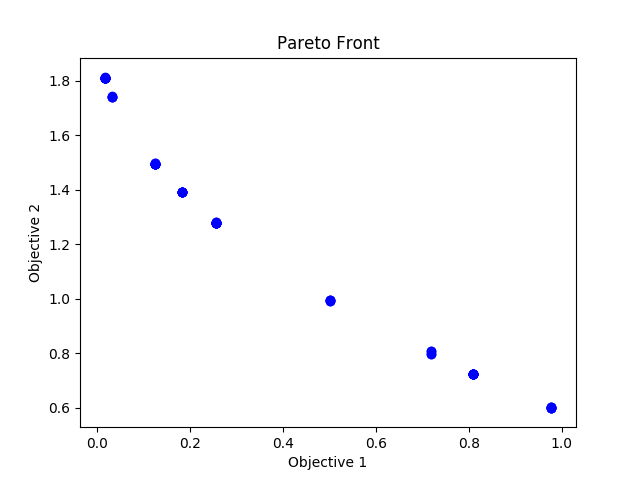}
\caption{Pareto Front obtained using original NSGA-II}
\label{fig:pareto_off}
\end{figure}

\begin{figure}[h!]
\centering
\includegraphics[width=0.5\textwidth]{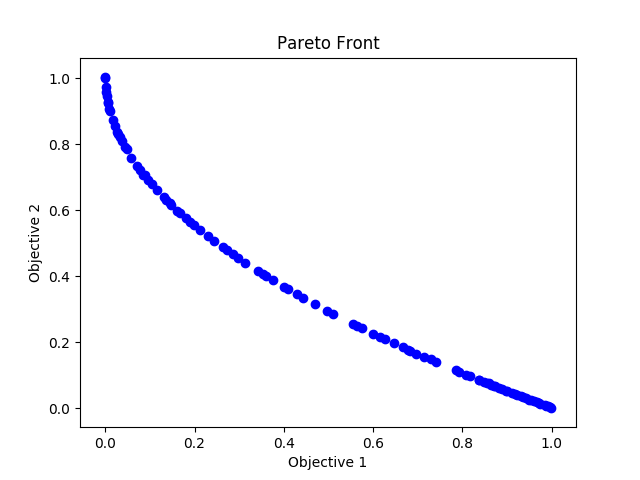}
\caption{Pareto Front obtained using CEC-NSGAII}
\label{fig:pareto_on}
\end{figure}

The results indicate that CEC-NSGAII algorithm converges faster and finds a better Pareto front compared to the original NSGA-II. The CEC-NSGAII algorithm achieves better distribution and diversity of solutions along the Pareto front, as evidenced by the more comprehensive coverage and denser clustering of solutions in Figure \ref{fig:pareto_on}.

\begin{figure}[h!]
\centering
\includegraphics[width=0.5\textwidth]{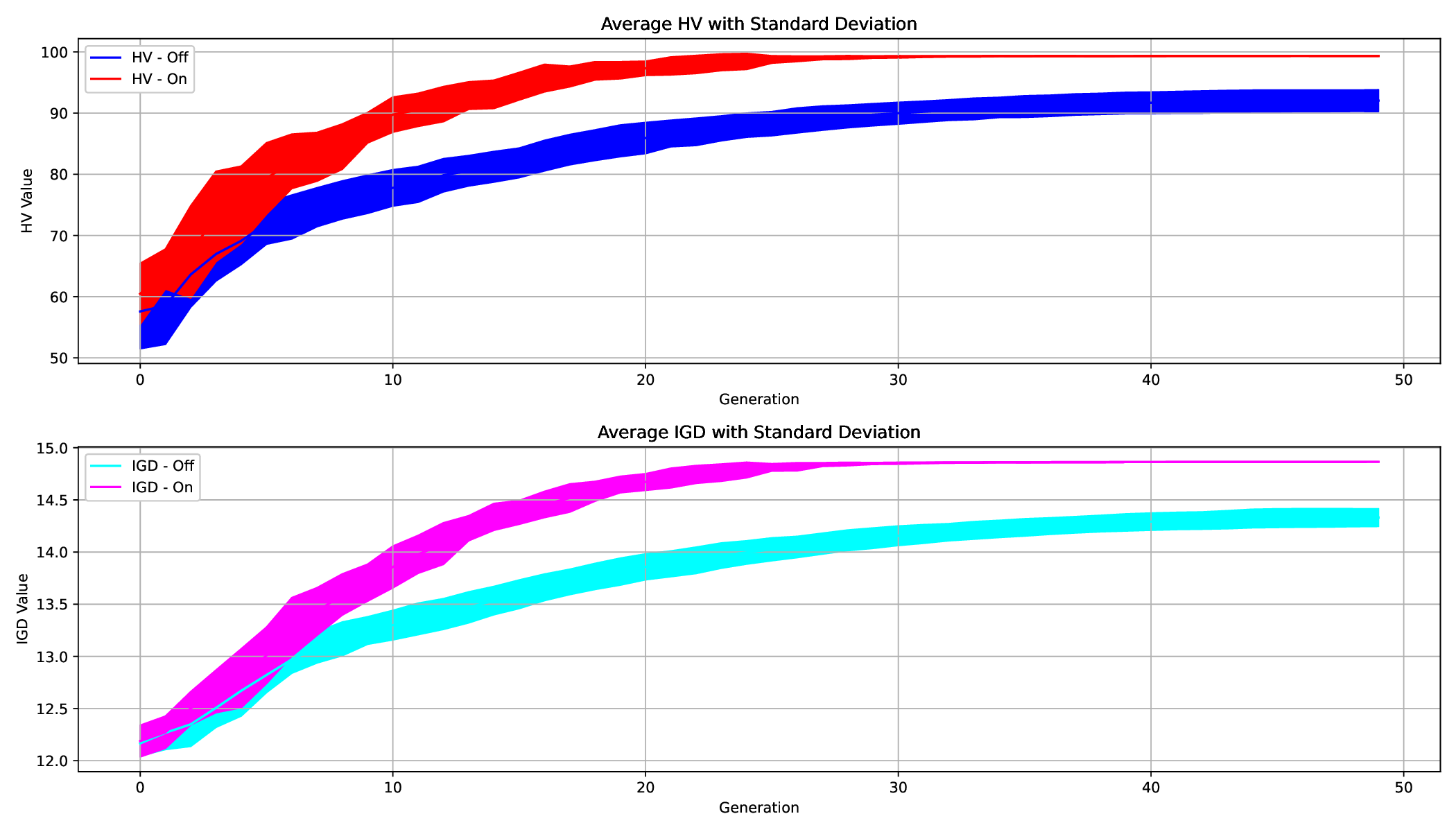}
\caption{HV and IGD values for ZDT1. The red and blue lines represent the HV values of our algorithm and the original NSGA-II, respectively. The magenta and cyan lines show the IGD values, demonstrating the superior performance of our algorithm in terms of convergence and diversity.}
\label{fig:matric_value_ZDT1}
\end{figure}

\begin{figure}[h!]
\centering
\includegraphics[width=0.5\textwidth]{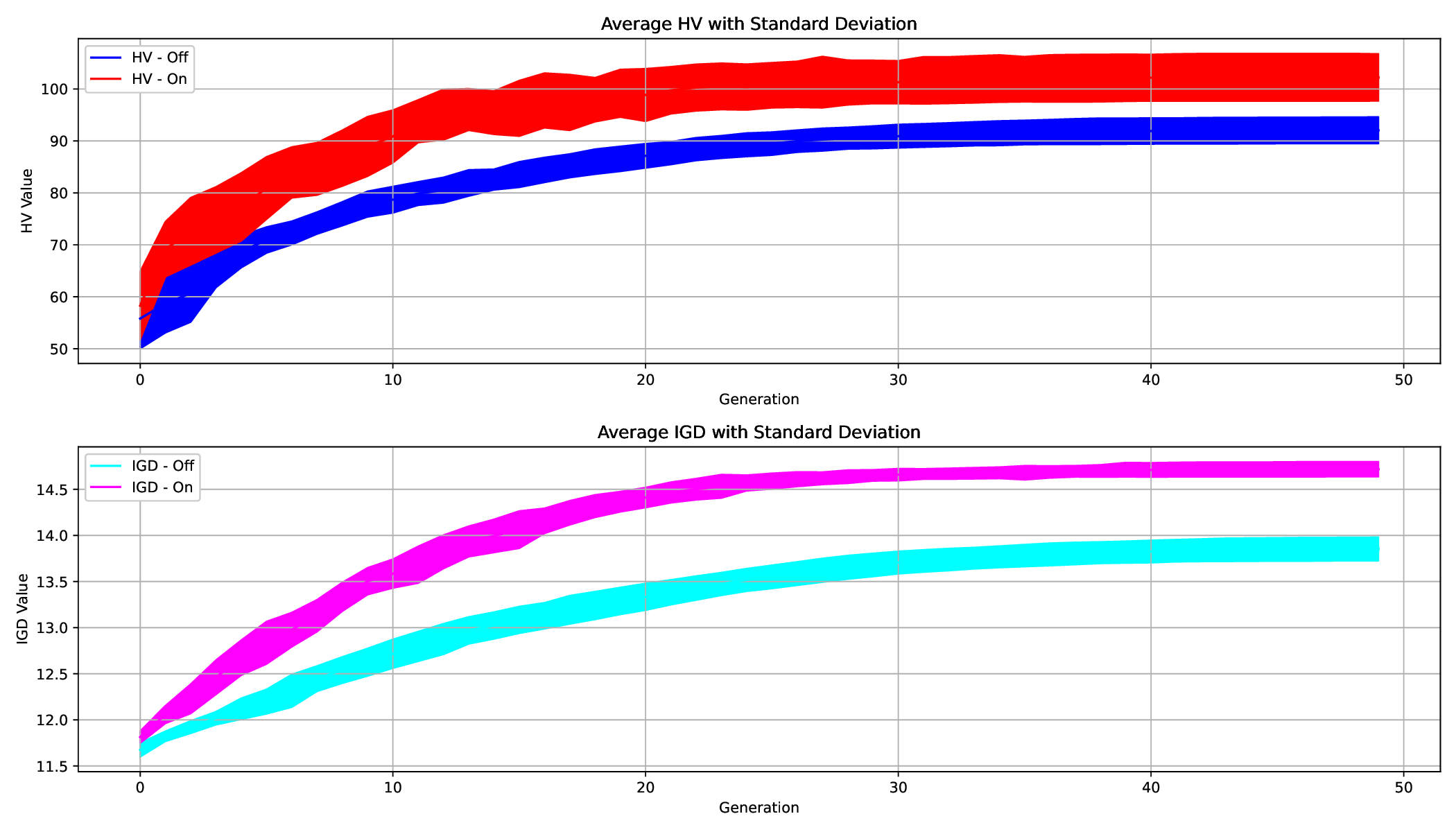}
\caption{HV and IGD values for ZDT2. The comparison shows that our algorithm consistently achieves better performance than the original NSGA-II in terms of both convergence and diversity.}
\label{fig:matric_value_ZDT2}
\end{figure}

\begin{figure}[h!]
\centering
\includegraphics[width=0.5\textwidth]{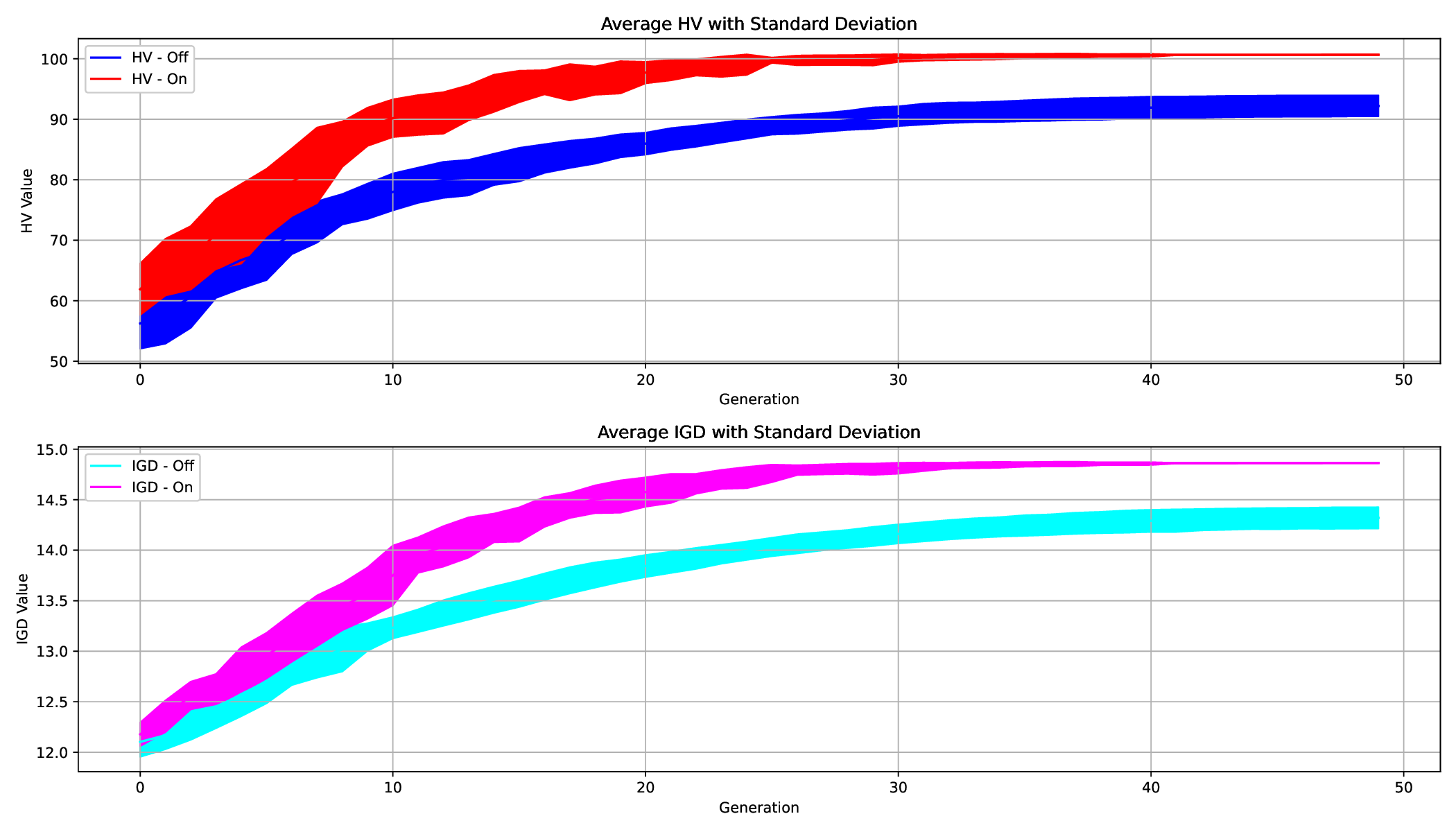}
\caption{HV and IGD values for ZDT3. The superior performance of our algorithm is highlighted by the higher HV and lower IGD values compared to the original NSGA-II.}
\label{fig:matric_value_ZDT3}
\end{figure}

Figures \ref{fig:matric_value_ZDT1}, \ref{fig:matric_value_ZDT2}, and \ref{fig:matric_value_ZDT3} present the comparison of our proposed algorithm with the original NSGA-II algorithm. These figures show the average performance over 21 experiments, highlighting the improvement in optimization metrics.

Figure \ref{fig:matric_value_ZDT1} displays the HV (Hypervolume) and IGD (Inverted Generational Distance) values for the ZDT1 benchmark. The red and blue lines represent the HV values of our algorithm and the original NSGA-II, respectively, while the magenta and cyan lines show the IGD values. The results demonstrate that our algorithm achieves higher HV and lower IGD values, indicating better performance in terms of both convergence and diversity.

Figure \ref{fig:matric_value_ZDT2} presents the HV and IGD values for the ZDT2 benchmark. Similar to the results for ZDT1, our algorithm outperforms the original NSGA-II, achieving higher HV and lower IGD values across all generations.

Figure \ref{fig:matric_value_ZDT3} illustrates the HV and IGD values for the ZDT3 benchmark. Once again, the results indicate that our algorithm significantly outperforms the original NSGA-II, as evidenced by the higher HV and lower IGD values throughout the optimization process.

\section{Conclusion}
\label{chap:Conclusion}

This paper presents innovative methodologies for tackling the complexities of SOMMOP and MOO by focusing on enhancing diversity and search efficiency in the search space, rather than the objective space. Unlike traditional approaches that primarily concentrate on objective space separation, the proposed methods delve into the search space itself to address the challenges posed by multi-modal landscapes.

For SOMMOP, we introduced the Chaotic Evolution with Deterministic Crowding (CEDC) and Chaotic Evolution with Clustering Algorithm (CECA). These algorithms leverage chaotic dynamics, persistence-based clustering, and Gaussian mutation to create a more effective search process that maintains high diversity while preventing premature convergence. Chaotic evolution introduces structured randomness, enabling the algorithms to escape local optima and thoroughly explore complex landscapes. Persistence-based clustering and enhanced niching techniques further support robust separation of peak regions in the search space, promoting effective discovery of multiple optima.

For MOO, our approach extends these techniques within a comprehensive multi-objective framework that incorporates Uncertainty-Based Selection and Adaptive Parameter Tuning. By introducing a radius \( R \) within deterministic crowding, the algorithm ensures precise separation of populations around peak solutions, maintaining solution diversity and stability in the search space. This radius concept aids in managing spatial relationships among solutions, particularly at convergence points, enabling a more stable and efficient exploration-exploitation balance.

Extensive experimental results on benchmark functions reveal that CEDC and CECA outperform conventional methods in terms of accuracy, robustness, and the ability to locate multiple optima. These findings underscore the value of search space-focused diversity management in multi-modal optimization and point towards new directions for evolutionary algorithms in handling complex, high-dimensional landscapes.

In future work, this approach could be further enriched by integrating machine learning-based fitness prediction models, enabling more precise adaptability in parameter tuning. Additionally, extending chaotic evolution principles to large-scale optimization and dynamic environments could yield further insights, particularly in high-dimensional and real-world applications where traditional methods remain limited. Overall, this work highlights a promising new pathway in optimization research, emphasizing the advantages of search space diversity and adaptive strategies in evolutionary algorithm design.

\bibliographystyle{IEEEtran}
\bibliography{main}

\begin{thebibliography}{10}
\providecommand{\url}[1]{#1}
\csname url@samestyle\endcsname
\providecommand{\newblock}{\relax}
\providecommand{\bibinfo}[2]{#2}
\providecommand{\BIBentrySTDinterwordspacing}{\spaceskip=0pt\relax}
\providecommand{\BIBentryALTinterwordstretchfactor}{4}
\providecommand{\BIBentryALTinterwordspacing}{\spaceskip=\fontdimen2\font plus
\BIBentryALTinterwordstretchfactor\fontdimen3\font minus
  \fontdimen4\font\relax}
\providecommand{\BIBforeignlanguage}[2]{{%
\expandafter\ifx\csname l@#1\endcsname\relax
\typeout{** WARNING: IEEEtran.bst: No hyphenation pattern has been}%
\typeout{** loaded for the language `#1'. Using the pattern for}%
\typeout{** the default language instead.}%
\else
\language=\csname l@#1\endcsname
\fi
#2}}
\providecommand{\BIBdecl}{\relax}
\BIBdecl

\bibitem{deb2002scalable}
K.~Deb, L.~Thiele, M.~Laumanns, and E.~Zitzler, ``Scalable multi-objective
  optimization test problems,'' in \emph{Proceedings of the 2002 Congress on
  Evolutionary Computation. CEC'02 (Cat. No. 02TH8600)}, vol.~1.\hskip 1em plus
  0.5em minus 0.4em\relax IEEE, 2002, pp. 825--830.

\bibitem{das2010differential}
S.~Das and P.~N. Suganthan, ``Differential evolution: A survey of the
  state-of-the-art,'' \emph{IEEE transactions on evolutionary computation},
  vol.~15, no.~1, pp. 4--31, 2010.

\bibitem{li2016seeking}
X.~Li, M.~G. Epitropakis, K.~Deb, and A.~Engelbrecht, ``Seeking multiple
  solutions: An updated survey on niching methods and their applications,''
  \emph{IEEE Transactions on Evolutionary Computation}, vol.~21, no.~4, pp.
  518--538, 2016.

\bibitem{eiben2003multimodal}
A.~Eiben, J.~Smith, A.~Eiben, and J.~Smith, ``Multimodal problems and spatial
  distribution,'' \emph{Introduction to Evolutionary Computing}, pp. 153--172,
  2003.

\bibitem{meng2022chaotic}
X.~Meng, Y.~Ding, and Y.~Pei, ``Chaotic evolution using deterministic crowding
  method for multi-modal optimization,'' in \emph{2022 IEEE International
  Conference on Systems, Man, and Cybernetics (SMC)}.\hskip 1em plus 0.5em
  minus 0.4em\relax IEEE, 2022, pp. 815--820.

\bibitem{meng2024evolutionary}
X.~Meng, Y.~Pei, and H.~Takagi, ``Evolutionary multi-modal optimization using
  persistence-based clustering in riemannian manifolds,'' in \emph{2024 IEEE
  Congress on Evolutionary Computation (CEC)}.\hskip 1em plus 0.5em minus
  0.4em\relax IEEE, 2024, pp. 1--8.

\bibitem{pei2014chaotic}
Y.~Pei, ``Chaotic evolution: fusion of chaotic ergodicity and evolutionary
  iteration for optimization,'' \emph{Natural Computing}, vol.~13, pp. 79--96,
  2014.

\bibitem{alatas2010chaotic}
B.~Alatas, ``Chaotic harmony search algorithms,'' \emph{Applied mathematics and
  computation}, vol. 216, no.~9, pp. 2687--2699, 2010.

\bibitem{caponetto2003chaotic}
R.~Caponetto, L.~Fortuna, S.~Fazzino, and M.~G. Xibilia, ``Chaotic sequences to
  improve the performance of evolutionary algorithms,'' \emph{IEEE transactions
  on evolutionary computation}, vol.~7, no.~3, pp. 289--304, 2003.

\bibitem{mahfoud1995niching}
S.~W. Mahfoud, \emph{Niching methods for genetic algorithms}.\hskip 1em plus
  0.5em minus 0.4em\relax University of Illinois at Urbana-Champaign, 1995.

\bibitem{edelsbrunner2022computational}
H.~Edelsbrunner and J.~L. Harer, \emph{Computational topology: an
  introduction}.\hskip 1em plus 0.5em minus 0.4em\relax American Mathematical
  Society, 2022.

\bibitem{mitchell1996introduction}
D.~Mitchell, ``Introduction: Public space and the city,'' \emph{Urban
  Geography}, vol.~17, no.~2, pp. 127--131, 1996.

\bibitem{li2013benchmark}
X.~Li, K.~Tang, M.~N. Omidvar, Z.~Yang, K.~Qin, and H.~China, ``Benchmark
  functions for the cec 2013 special session and competition on large-scale
  global optimization,'' \emph{gene}, vol.~7, no.~33, p.~8, 2013.

\bibitem{coello2007evolutionary}
C.~A.~C. Coello, \emph{Evolutionary algorithms for solving multi-objective
  problems}.\hskip 1em plus 0.5em minus 0.4em\relax Springer, 2007.

\bibitem{zitzler2001spea2}
E.~Zitzler, M.~Laumanns, and L.~Thiele, ``Spea2: Improving the strength pareto
  evolutionary algorithm,'' \emph{TIK report}, vol. 103, 2001.

\bibitem{beume2007sms}
N.~Beume, B.~Naujoks, and M.~Emmerich, ``Sms-emoa: Multiobjective selection
  based on dominated hypervolume,'' \emph{European Journal of Operational
  Research}, vol. 181, no.~3, pp. 1653--1669, 2007.

\bibitem{zhang2007moea}
Q.~Zhang and H.~Li, ``Moea/d: A multiobjective evolutionary algorithm based on
  decomposition,'' \emph{IEEE Transactions on evolutionary computation},
  vol.~11, no.~6, pp. 712--731, 2007.

\bibitem{deb2013evolutionary}
K.~Deb and H.~Jain, ``An evolutionary many-objective optimization algorithm
  using reference-point-based nondominated sorting approach, part i: solving
  problems with box constraints,'' \emph{IEEE transactions on evolutionary
  computation}, vol.~18, no.~4, pp. 577--601, 2013.

\bibitem{deb2011multi}
K.~Deb, ``Multi-objective optimisation using evolutionary algorithms: an
  introduction,'' in \emph{Multi-objective evolutionary optimisation for
  product design and manufacturing}.\hskip 1em plus 0.5em minus 0.4em\relax
  Springer, 2011, pp. 3--34.

\bibitem{jin2011surrogate}
Y.~Jin, ``Surrogate-assisted evolutionary computation: Recent advances and
  future challenges,'' \emph{Swarm and Evolutionary Computation}, vol.~1,
  no.~2, pp. 61--70, 2011.

\bibitem{knowles2006parego}
J.~Knowles, ``Parego: A hybrid algorithm with on-line landscape approximation
  for expensive multiobjective optimization problems,'' \emph{IEEE transactions
  on evolutionary computation}, vol.~10, no.~1, pp. 50--66, 2006.

\bibitem{deb2002fast}
K.~Deb, A.~Pratap, S.~Agarwal, and T.~Meyarivan, ``A fast and elitist
  multiobjective genetic algorithm: Nsga-ii,'' \emph{IEEE transactions on
  evolutionary computation}, vol.~6, no.~2, pp. 182--197, 2002.

\bibitem{mengshoel1999probabilistic}
O.~J. Mengshoel and D.~E. Goldberg, ``Probabilistic crowding: Deterministic
  crowding with probabilistic replacement,'' 1999.

\bibitem{zitzler2000comparison}
E.~Zitzler, K.~Deb, and L.~Thiele, ``Comparison of multiobjective evolutionary
  algorithms: Empirical results,'' \emph{Evolutionary computation}, vol.~8,
  no.~2, pp. 173--195, 2000.

\bibitem{huband2005scalable}
S.~Huband, L.~Barone, L.~While, and P.~Hingston, ``A scalable multi-objective
  test problem toolkit,'' in \emph{Evolutionary Multi-Criterion Optimization:
  Third International Conference, EMO 2005, Guanajuato, Mexico, March 9-11,
  2005. Proceedings 3}.\hskip 1em plus 0.5em minus 0.4em\relax Springer, 2005,
  pp. 280--295.

\bibitem{deb2006multi}
K.~Deb, A.~Sinha, and S.~Kukkonen, ``Multi-objective test problems, linkages,
  and evolutionary methodologies,'' in \emph{Proceedings of the 8th annual
  conference on Genetic and evolutionary computation}, 2006, pp. 1141--1148.

\bibitem{sareni1998fitness}
B.~Sareni and L.~Krahenbuhl, ``Fitness sharing and niching methods revisited,''
  \emph{IEEE transactions on Evolutionary Computation}, vol.~2, no.~3, pp.
  97--106, 1998.

\end{thebibliography}

\end{document}